\documentclass[conference]{IEEEtran}

\IEEEoverridecommandlockouts
\usepackage{cite}
\usepackage{amsmath,amssymb,amsfonts,subfigure,enumitem,multirow}
\usepackage{algorithmic}
\usepackage{graphicx}
\usepackage[ruled,vlined]{algorithm2e}
\usepackage{textcomp}
\usepackage{dsfont}
\usepackage{xcolor}
\usepackage{caption}
\usepackage{booktabs}
\usepackage{url}
\usepackage{makecell}
\usepackage{footnote}
\def\BibTeX{{\rm B\kern-.05em{\sc i\kern-.025em b}\kern-.08em
    T\kern-.1667em\lower.7ex\hbox{E}\kern-.125emX}}
\begin{document}
\makesavenoteenv{table}

\title{UCTB: An Urban Computing Tool Box for Building Spatiotemporal Prediction Services}

\author{\IEEEauthorblockN{Jiangyi Fang$^{1,5*}$, Liyue Chen$^{1,5*}$, Di Chai$^{2*}$, Yayao Hong$^{3,4}$, Xiuhuai Xie$^{3,4}$,Longbiao Chen$^{3,4}$,Leye Wang$^{1,5\dagger }$}
\IEEEauthorblockA{$^{1}$ \textit{Key Lab of High Confidence Software Technologies}, \textit{Ministry of Education, China}}
\IEEEauthorblockA{$^{2}$ \textit{Hong Kong University of Science and Technology, Hong Kong}}

\IEEEauthorblockA{$^{3}$ \textit{Fujian Key Laboratory of Sensing and Computing for Smart Cities, Xiamen}}
\IEEEauthorblockA{$^{4}$ \textit{School of Informatics, \textit{Xiamen University}, Xiamen}\\}
\IEEEauthorblockA{$^{5}$ \textit{School of Computer Science, \textit{Peking University}, Beijing}\\
\{fangjiangyi2001,chenliyue2019\}@gmail.com,dchai@cse.ust.hk,leyewang@pku.edu.cn }
\thanks{$^*$ All authors contributed equally to this research. D. Chai developed the first version of UCTB in 2018-2019. L. Chen has been a leading developer and maintainer of UCTB since 2020. J. Fang has been a leading developer and maintainer of UCTB since 2023.}
\thanks{$^\dagger$ Corresponding author}
}

\maketitle
\vspace{-2em}
\begin{abstract}
Spatiotemporal prediction (STP) service is one of the key infrastructure applications in smart cities. Currently, most of the existing STP services are constructed following the workflow of building deep learning (DL) applications while neglecting the importance of domain knowledge and region partition. However, the performance and interpretability of STP are highly related to them. As a result, there is an urgent requirement to develop a thorough and tailored workflow for STP services.
To address this gap, we propose a novel workflow including two factors above as intermediate procedures. Based on the workflow, we design and implement an STP toolbox called \textbf{UCTB} (Urban Computing Tool Box) assisting practitioners in the rapid construction of STP services, which can manage multiple spatiotemporal domain knowledge, support various region partition algorithms, and possess state-of-the-art models simultaneously. The relevant code and supporting documents have been open-sourced at https://github.com/uctb/UCTB.
\end{abstract}


\section{Introduction} \label{sec: intro}
With the advancement of sensor networks, mobile intelligent terminals, and location acquisition technologies~\cite{chen2019review}, vast amounts of data containing time and geographic information have become available. Such data includes vehicle speed, supply-demand intensity, and traffic flow - collectively known as spatiotemporal data~\cite{yuan2021survey}. These rich spatiotemporal datasets provide an excellent foundation for AI-based technology, which is a key sign of smart city~\cite{zheng2014urban}. As a result, many researchers make great efforts to develop advanced learning-based technologies for many tasks in urban computing~\cite{ASTGCN_2019,graphwavenet_2019,mtgnn,attention_2017,jin2022deep,ji2022stden,wang2021gallat,zhang2021traffic,shao2022decoupled,li2023dynamic,lan2022dstagnn,fang2021spatial}. As a crucial aspect of urban computing, Spatio-Temporal Prediction (STP) provides fundamental services for numerous applications in the transport domain such as traffic control~\cite{lin2023predicting}, order dispatching, site selection, and route planning~\cite {yuan2021survey}. Therefore, it poses a large demand for practitioners to adapt advanced technologies to specific scenarios and construct corresponding STP services.


While most existing STP services are developed following a typical (DL) workflow ~\cite{miao2017towards,wang2020deep,baltensperger2022continuous} including Data Preparation, Model Definition, Training and Evaluation, STP services have unique characteristics that necessitate a tailored workflow for them. For example, STP services require region partitioning as a prerequisite, which differs from the typical DL workflow but significantly impacts the services. Recent studies have revealed that an appropriate region partition customized for specific tasks greatly affects both the interpretability and effectiveness of STP services~\cite{chen2023data,jin2022gridtuner}. Neglecting these influencing factors when building STP services may lead to sub-optimal performance and inexplainable results~\cite{wang2020deep,jin2023spatio}.  Except for region partition, practitioners can improve the performance of STP services through the integration of domain knowledge. Take temporal knowledge as an example. Prior studies have demonstrated improved prediction accuracy by incorporating historical traffic patterns with considerations for closeness, daily periodicity, and weekly periodicity~\cite{zhang2017deep,deepstn_2019, ASTGCN_2019}. In the conventional DL workflow, leveraging domain knowledge from experts is commonly integrated into the process of model definition~\cite{xie2021survey,}. Indeed, given the volatile attributes of scenarios, the careful selection of domain knowledge may prove more efficacious than the development of fancy model architectures~\cite{STMeta}. Therefore, practitioners must regard domain knowledge as an integral component separate from modeling techniques. \textbf{In summary, there is an urgent need to establish a comprehensive and tailored workflow for STP services}.

However, most existing STP studies overlook the significance of domain knowledge and region partition, thereby hindering STP services from meeting various requirements of scenarios. To address these limitations, we propose a workflow tailored to the distinctive features of constructing STP services. Specifically, we establish the following two fundamental design principles:
 \begin{enumerate}
     \item \textbf{Generalization}: The workflow provides a unified descriptive framework for every STP service specialized for the corresponding scenario.
     \item \textbf{Extensibility}: Faced with novel scenarios, the workflow should possess the capacity to incorporate an extensive set of domain knowledge and emerging technologies.
 \end{enumerate}
As an example to show the unified character, we set up a procedure termed ``Region Generation'' as one of the prerequisites of the STP model definition. In order to implement different methods of region partition in a unified way, the ``Region Generation'' procedure is divided into three steps: region partition, spatial binding and spatial aggregation. Besides, We stipulate that the inputs and outputs of this procedure should have formats decoupled to specific methods to support the user's customized implementations for flexible extensibility.
To help developers build STP services with our workflow, we have developed a toolbox called UCTB (Urban Computing Tool Box), aimed at facilitating the creation of spatiotemporal prediction services aligned with the proposed workflow. UCTB includes specific modules for each procedure of the workflow, aiding in the development of STP services from scratch. We encapsulate our contributions in the following points:

\begin{itemize}
    \item We propose a novel workflow for constructing STP service with region generation and knowledge management, which is closer to the demands (more flexible spatial units and generalization to new scenarios) of the real scenarios.

    \item We design and implement a toolbox named UCTB based on our proposed workflow to assist practitioners in quickly building and testing STP services. This toolbox contains numerous statistical and DL models and an extensible framework with various reusable layers for users to implement customized STP services.

    \item We conduct extensive experiments through the UCTB toolbox to justify the effectiveness of our proposed workflow. Besides, we display a thorough use case of UCTB in a given scenario.

    \item We have established a standardized data format for spatiotemporal data that seamlessly integrates with UCTB, enabling researchers and practitioners to efficiently test their services using their own data. Additionally, we have developed a visualization tool specifically designed to identify and diagnose errors within the results of STP services.
\end{itemize}

\section{Related Work}

Recently, we have witnessed many efforts toward building effective and generalized spatiotemporal prediction services. In these pioneer works, both open-source datasets and services benefit the research community a lot. We list some popular datasets related to STP in table~\ref{tab: dataset_comparison}.

\begin{table}[htbp]
  \small
  \caption{Datasets Related to STP.}
  \label{tab: dataset_comparison}
  \centering
  \begin{tabular}{lccccccccc}
    \toprule
    \textbf{Dataset} & \textbf{Type}  \\
    \midrule
    \textbf{\textit{Node-based Spatiotemporal Dataset}} \\
    METR-LA, PEMS-BAY~\cite{li2018dcrnn_traffic} & Highway Speed \\
    PEMSD3, PEMSD7~\cite{STSGCN_2020} & Highway Speed  \\
    PEMSD4, PEMSD8~\cite{ASTGCN_2019} & Highway Speed  \\ 
    PEMS7(L), PEMS7(M)~\cite{stgcn_2018}  & Highway Speed \\
    Melbourne Pedestrian\footnote{https://data.melbourne.vic.gov.au/explore/dataset/pedestrian-counting-system-monthly-counts-per-hour/information/} & Pedestrian Count\\
    \midrule
    \textbf{\textit{Grid-based Spatiotemporal Dataset}} \\
    NYCTaxi~\cite{yao_similarity_2019} & Taxi Trip \\
    NYCBike~\cite{yao_similarity_2019} & Bike Trip \\
    TaxiBJ2014, TaxiBJ2015~\cite{zhang2017deep} & Taxi Trip\\
    \midrule
    \textbf{\textit{External Dataset}} \\
    Foursquare NYC and Tokyo ~\cite{yang2014modeling} & Foursquare Check-in \\
    Gowalla~\cite{friendship_2011} & Friendship Network \\
    \bottomrule
  \end{tabular}
\end{table}

At the same time, researchers have constructed many STP services and released their codes including \textit{ST-ResNet}~\cite{zhang2017deep}, \textit{DCRNN}~\cite{li2018dcrnn_traffic}, \textit{STGCN}~\cite{stgcn_2018}, GraphWaveNet~\cite{graphwavenet_2019}, \textit{GMAN}~\cite{GMAN_2020}, and \textit{STMeta}~\cite{STMeta}. A popular GitHub repository\footnote{https://github.com/aptx1231/Traffic-Prediction-Open-Code-Summary} has summarized various open-source STP services for traffic prediction. Please be aware that while these works offer open-source codes for STP, it is crucial to carefully review the documentation and follow the instructions provided in order to understand their specific usage and adapt them to meet target scenarios. Additionally, the workflows of these prediction models vary (e.g., feature transformation and normalization techniques), which makes it difficult to make a fair comparison between existing STP services.

\textit{LibCity} is an open-source library that offers a wide range of tools, datasets, and models for urban computing tasks~\cite{libcity_2021,wang2023towards}. The primary objective of \textit{LibCity} is to simplify the application and evaluation of algorithms related to urban data analysis, traffic prediction, and intelligent transportation systems. The toolbox includes various STP models such as time series models and graph neural networks. These models can be utilized to construct STP services. 

Although both UCTB and \textit{Libcity} share overarching objectives, UCTB distinguishes itself as a toolbox founded on our custom-designed workflow tailored for STP. Unlike \textit{Libcity}, UCTB incorporates specialized modules aimed at facilitating region generation and knowledge management within the spatiotemporal prediction workflow. For example, function \textit{partition} implements various regional divisions, function \textit{binding} implements binding between units and regions, Class \textit{ST\_MoveSample} samples based on temporal knowledge to obtain temporal features, and Class \textit{GraphGenerator} builds graphs based on spatial knowledge to capture spatial correlations. Moreover, UCTB prioritizes extensibility and code reusability by introducing a \textit{BaseModel} as the foundational class for other models and incorporating numerous reusable layers in the model definition phase, streamlining the development of spatiotemporal prediction services tailored to specific scenario requirements.

In other words, given a specific STP task, rather than directly using existing STP services, developers can use UCTB to easily construct services based on their own needs (e.g., creating spatial graphs useful for the target task, even if these graphs are not used in the model's original paper).
Specifically, UCTB provides both spatial and temporal feature transformation interfaces (i.e., \textit{ST\_MoveSample} and \textit{GraphGenerator}) which may help developers to flexibly generate useful temporal sequence features and spatial graphs for the target prediction task.

\section{Workflow Methodology}

\begin{figure*}[t]
  \centering
  \includegraphics[width=.88\linewidth]{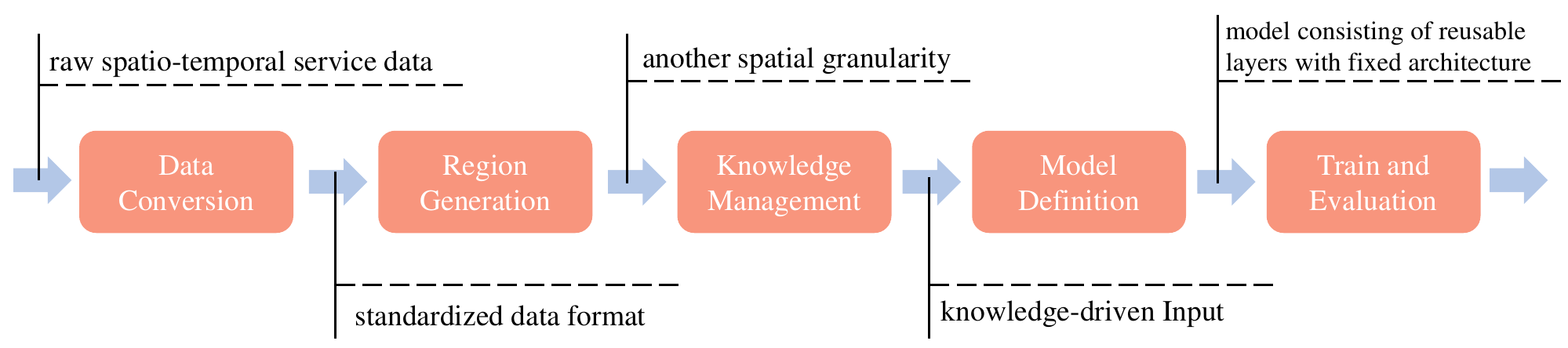}
  \caption{Proposed Workflow.}
  \label{fig:workflow}
  \vspace{-1em}
\end{figure*}

\subsection{Workflow Overview}
The workflow we propose generally contains five procedures: (1) data conversion, (2) region generation, (3) knowledge management, (4) model definition, and (5) training and evaluation as shown in Figure~\ref{fig:workflow}. 

Within ``data conversion'', a standardized dataset format should be defined to apply to various STP scenarios. Based on the standardized format, ``region generation'' functions as changing the spatial granularity of the spatiotemporal dataset to obtain regions with better characteristics. During ``knowledge management'', two consecutive steps of knowledge selection and feature transformation are conducted to leverage spatiotemporal knowledge for better prediction results.
Within the model definition process, we construct reusable advanced layers and formulate prediction models, encompassing both existing methodologies and newly devised ones, leveraging these layers as building blocks.
Finally, during training and evaluation, one should implement strategies for both model training and evaluation. The training strategy will focus on optimizing the model parameters to improve prediction accuracy. The evaluation strategy will assess the performance of service based on certain metrics using separate test datasets.

\subsection{Data Conversion}

The procedure of data Conversion not only presents in existing works of STP library~\cite{shao2023exploring,libcity_2021}, but it is also an fundamental part in time series prediction librarys\footnote{https://github.com/jgasthaus/gluon-ts}. This procedure often defines a standardized data format as an intermediary between raw data and the interface that subsequent procedures can accept. Users only need to preprocess their raw data into this format before using UCTB for their scenarios.


\subsection{Region Generation}
As an instance of the widely known problem the \textit{modefiable areal unit problem} (MAUP)~\cite{wong2004modifiable,openshaw1981modifiable}, how to determine the regions (spatial units) for STP service is of great importance, since most different region specifications may lead to diverged analysis outcomes. For the procedure of region generation, we believe that it contains three key steps: partition, binding and aggregation. The input of the procedure should be the specific data format defined in the data conversion part. The output should be in the same format as the input but with another spatial granularity and specification.
\subsection{Knowledge Management}
Basically, there are two main steps in this procedure, knowledge selection and feature transformation. The step of knowledge selection is responsible for determining which factors should be incorporated when building STP service in the specific scenario, while the feature transformation functions as a tool to Convert spatiotemporal traffic data and other related information into corresponding forms of input based on knowledge selection results.
\subsubsection{Knowledge Selection}
Generalized factors affecting STP can be classified into three parts: temporal knowledge, spatial knowledge and external factors.

\begin{figure*}[t]
  \centering
  \includegraphics[width=.7\linewidth]{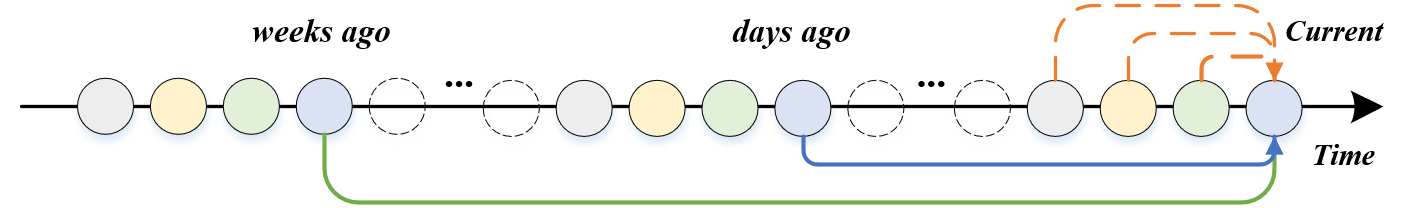}
  \caption{Temporal Transformation: Generating Temporal Features by Sampling Time Series Data. }
  \label{fig: st_move}
  \vspace{-1em}
\end{figure*}

\textbf{Temporal knowledge} plays a crucial role in this feature transformation process by projecting future traffic values based on past traffic values sampled at different time intervals (refer to Figure~\ref{fig: st_move}). There are three common types of temporal knowledge:
\begin{enumerate}
    \item \textit{Closeness}: Temporal data at adjacent moments typically do not experience substantial changes, implying that past traffic values are related to future predicted values. 
    \item \textit{Daily Periodicity}: Future traffic values often correspond to the same moment on previous days, reflecting the daily periodicity of human activities.
    \item \textit{Weekly Periodicity}: The flow on Saturday resembles that on the previous Saturday compared to other weekdays.
\end{enumerate}

\textbf{Spatial knowledge}, which reflects the relationship between different locations, has been extensively studied in previous research \cite{Zhang2016DNNbasedPM, STMeta}. There are mainly three common types of spatial knowledge.

\begin{enumerate}
    \item \textit{Proximity}: Units that are close to each other tend to have higher correlations.
    \item \textit{Functionality}: Time series between similar functional areas may enhance each other's predictions
    \item \textit{Inter-Location Relationship}: Locations which possess the same attribute may have higher correlations e.g. speed sensors on the same road.
\end{enumerate}

Apart from spatiotemporal features, \textbf{external factors} like weather conditions affect patterns of spatiotemporal data. For example, heavy rain and strong winds can reduce taxi demand \cite{li_traffic_2015}. As a result, the knowledge selection procedure is also responsible for choosing which external factors to incorporate and afterwards processing of these external variables.
\subsubsection{Feature Transformation}
Feature transformation utilizes selected prior knowledge to convert spatiotemporal traffic data and other information into various kinds of input. For every temporal knowledge, we sample several times at corresponding time slots to form a input sequence~\cite{zhang2017deep}. For every spatial knowledge, we construct a graph to represent the relationship between spatial units. This process helps predictive models capture different patterns of traffic flow under different spatial relationships more effectively. 
\subsection{Model Definition}
There are two main categories of prediction models for STP services, namely statistical learning models and deep learning models. Although these types of models differ in construction and training processes, it is convenient for the user if the specific prototype encapsulates them and provides similar user interfaces. 

Besides, when implementing DL models, some training and prediction interfaces as well as model storage interfaces are similar. To maximize code reuse, we need to design a basic model class that includes functions such as training, prediction, breakpoint continuation training, etc. Therefore, specific DL models only need to inherit this basic class while defining their model structure and setting corresponding feature input functions.

\subsection{Training and Evaluation}
Training statistical learning models is relatively straightforward. When it comes to training DL models, it's common practice to divide the data into batches and select small batches for gradient updates during training. This process needs to be integrated into the specific toolbox in a reusable way. Additionally, early stopping mechanisms are often used as they help train converged models with fewer epochs.

Once model training and testing are finished, prediction results must undergo appropriate evaluation by comparing actual values with predicted ones. In STP problems, RMSE (Root Mean Square Error), MAE (Mean Absolute Error), and MAPE (Mean Absolute Percentage Error) are widely used for evaluation purposes. Therefore, corresponding evaluation interfaces need to be implemented.
\begin{figure*}[t]
  \centering
  \includegraphics[width=.85\linewidth]{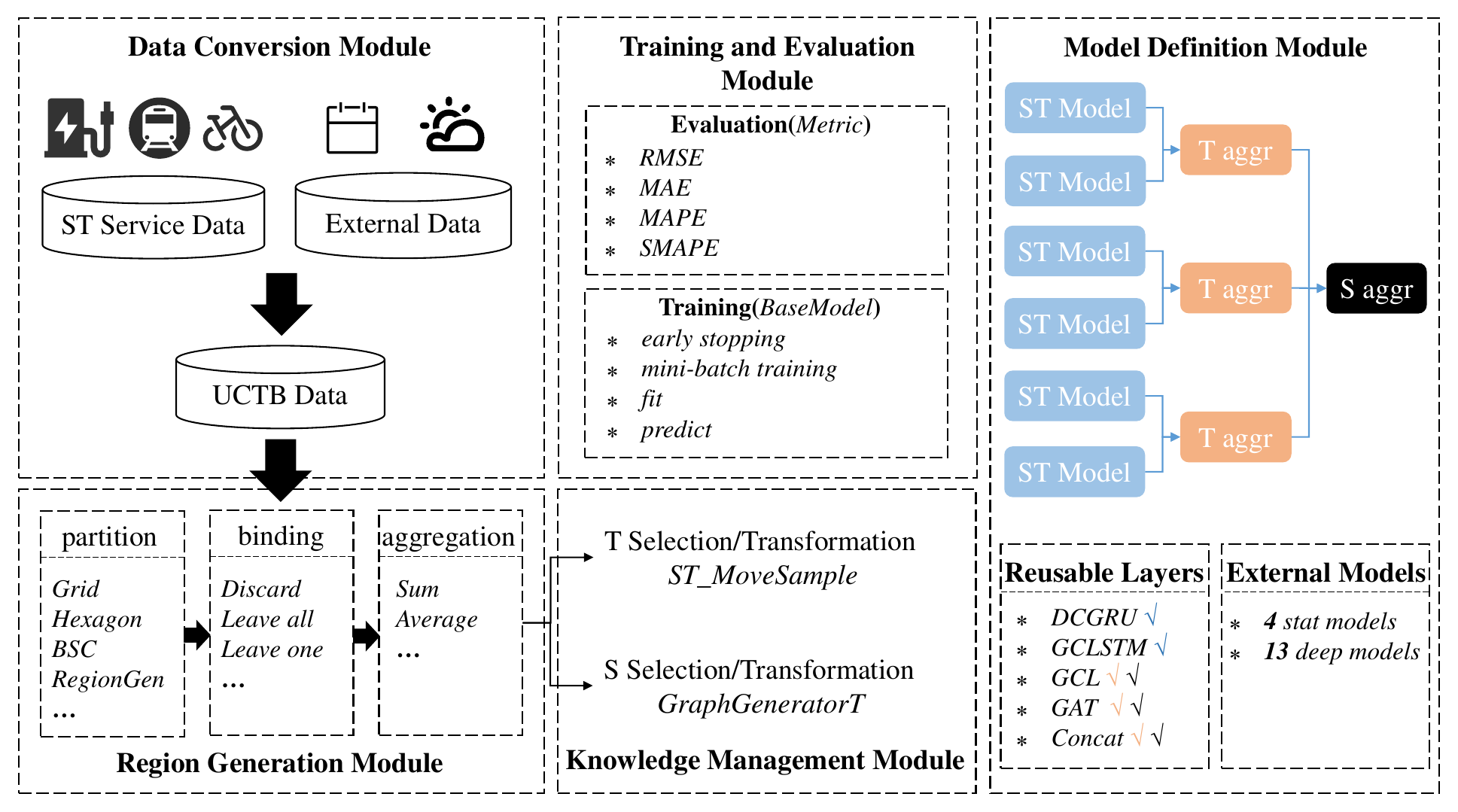}
  \vspace{-1em}
  \caption{Overview of UCTB Toolbox.}
  \label{fig:framework}
  \vspace{-1em}
\end{figure*}
\section{Implementation and Interface}
The overview of UCTB toolbox is shown in Figure~\ref{fig:framework}, where each functional module corresponds to a step in the above workflow. In this section, we will introduce the implementation and the interfaces of UCTB. We mainly focus on five major modules including data conversion, region generation, knowledge management, model definition, and training \& evaluation. 

\subsection{Data Conversion Interface}
The UCTB standardized dataset is structured as key-value pairs using the pickle protocol. The primary information for each key-value pair is listed in Table~\ref{table: dataset_format}. Additionally, in order to convert raw data conveniently, we implement a corresponding interface \textit{build\_uctb\_dataset} to help users quickly organize data into standardized datasets.

\begin{table}[h]
    \centering
    \caption{UCTB Dataset Format.}
    \begin{tabular}{ccccccc}
        \toprule
         \textbf{Key} & \textbf{Description}  \\ 
         \midrule
         TimeRange & The time range of the dataset  \\
         TimeFitness & The time interval of data  \\
         Node & 2-D (node) traffic flow data \\
         Grid & 3-D (grid) traffic flow data  \\
         ExternalFeature & Store external features  \\
        \bottomrule
    \end{tabular}
    \label{table: dataset_format}
\end{table}
\subsection{Region Generation Interface}
We summarize the interface of the Region Generation module in Table \ref{table: region_gen}. The \textit{partition} interface takes the boundary of the city, the method of partition, traffic flow data (optional) as input and outputs smaller spatial regions of the given city. It contains mainly two types of methods, prior and traffic data-driven as shown in Table~\ref{table: rp_methods}. For prior methods of region partition, the city will be partitioned into regular/irregular regions without any information of traffic data, while as to data-driven methods, regions will benefit from traffic data to be expected to attain better performance on downstream prediction tasks.

\begin{table}[h]
\centering
\caption{Region Generation Interface.}
\begin{tabular}{ccccccc}
\toprule
\textbf{Interface} & \textbf{Description} \\ 
\midrule
\textit{partition} & Partition a large spatial area in to smaller ones \\
\textit{binding} & Construct a mapping from units to regions \\
\textit{aggregation} & Aggregate time series according to the mapping \\
\bottomrule
\end{tabular}
\label{table: region_gen}
\end{table}

The \textit{binding} interface takes original spatial units of UCTB data and regions as input and outputs mapping between original spatial units and generated regions. The mapping is mainly determined by the spatial relationship between original spatial units and generated regions. The interface should be robust enough to handle the situation that there are some spatial units with no regions spatially contained.

\begin{table}[h]
    \centering
    \caption{Methods of Region Partition.}
    \begin{tabular}{ccccccc}
        \toprule
         \textbf{Method} & \textbf{Reference}  \\ 
         \midrule
         \textbf{Prior Methods}&& \\
         \textit{Grid} & \cite{zhang2017deep}  \\
         \textit{Hexagon} & \cite{sthan_2022}  \\
         \textit{Road Segment} & \cite{yuan2012segmentation} \\
         \midrule
         \textbf{Data-driven Methods}&& \\
         \textit{Greedy} & \cite{chen2023data}  \\
         \textit{D-balanced} & \cite{chen2023data}  \\
         \textit{Fluid} & \cite{pares2017fluid}  \\
         \textit{BSC} & \cite{li2015traffic}  \\
         \textit{GCSC} & \cite{chen2016dynamic}  \\
         \textit{RegionGen} & \cite{chen2023data}  \\
        \bottomrule
    \end{tabular}
    \vspace{-2em}
    \label{table: rp_methods}
\end{table}

The \textit{aggregation} interface takes the mapping and original 2-D traffic data as input and outputs spatially aggregated 2-D traffic data. We implement three methods of aggregation function:\textit{sum},\textit{average} for possible scenarios of taxi demand and traffic speed. Besides, we also support the customized aggregation function as an optional input.

\subsection{Knowledge Management Interface}
We summarize the interface of the knowledge management module in Table~\ref{table: feature_inter}.
The \textit{GridTrafficLoader} and \textit{NodeTrafficLoader} interfaces are used to load Euclidean and non-Euclidean data, respectively. These data loaders transform the raw traffic flow data into features based on spatiotemporal prior knowledge. The \textit{ST\_MoveSample} interface samples traffic data at different time intervals to obtain temporal features, while the \textit{GraphGenerator} interface generates various spatial graphs to obtain spatial features. Users can also inherit from the \textit{GraphGenerator} interface and implement new graphs for their scenarios.

\begin{table}[h]
\centering
\caption{Knowledge Management Interface.}
\begin{tabular}{ccccccc}
\toprule
\textbf{Interface} & \textbf{Description} \\ 
\midrule
\textit{GridTrafficLoader} & Load and preprocess grid data \\
\textit{NodeTrafficLoader} & Load and preprocess node data \\
\textit{ST\_MoveSample} & Sample features at different intervals \\
\textit{GraphGenerator} & Generate different types of graphs \\
\bottomrule
\end{tabular}
\label{table: feature_inter}
\end{table}

\textit{ST\_MoveSample} has three main sampling intervals that represent the number of sampled time features with different interval times. For example, adjacency similarity is 6; daily similarity is 7; weekly similarity is 4. This means that flow at six previous moments before prediction time, along with flow at the same moment during the last seven days and four weeks, are jointly considered as temporal features for prediction.

\textit{GraphGenerator} receives a graph name and generates its adjacency matrix and Laplacian matrix based on a corresponding threshold value. For instance, for a distance graph, we can set the threshold parameter to 6,500 meters. \textit{GraphGenerator} will generate an adjacency matrix based on Euclidean distances between each unit in which unitss less than or equal to this threshold are associated (set as 1) while those greater than it are not associated (set as 0). Selecting an appropriate threshold determines whether spatial knowledge can be well extracted or not. Based on our experimental experience, a better threshold generally allows each node to have connections with approximately 20\% of other nodes' average connections.

\vspace{-0.5em}

\subsection{Model Definition Interface}
UCTB offers two types of model interfaces: complete predictive models and reusable layers that are widely used across different STP scenarios. The latter enables users to easily create new custom models.

\begin{table}[h]
\centering
    \caption{Prediction Models in UCTB. (\textbf{T. Tech.} for Temporal Modeling Technology; \textbf{T. Know.} for Temporal Knowledge; \textbf{S. Tech.} for Spatial Modeling Technology; Attn for attention mechanism; C for Closeness; D for Daily Periodicity; W for Weekly Periodicity; P for Prior; A for Adaptive)}
    \begin{tabular}{lcccc}
        \toprule
         \textbf{Model} & \textbf{T. Tech.} & \textbf{T. Know.}& \textbf{S. Tech.}& \textbf{Graph}\\ 
         \midrule
         \textbf{statistical models}&& \\
         \textit{ARIMA}~\cite{1995Short} & SARIMA & C & N/A & N/A \\
         \textit{HM}  & N/A & C & N/A & N/A   \\
         \textit{HMM} & HMM & C & N/A & N/A   \\
         \textit{XGBoost}  & XGBoost & C & N/A & N/A    \\
            \midrule
         \textbf{deep models}&& \\
         \textit{DeepST}~\cite{Zhang2016DNNbasedPM}  & CNN & C-D-W & CNN & N/A  \\
         \textit{ST-ResNet}~\cite{zhang2017deep}  & CNN & C-D-W & CNN & N/A  \\
         \textit{DCRNN}~\cite{li2018dcrnn_traffic} & RNN & C & GNN & P \\
         \textit{GeoMAN}~\cite{GeoMAN_2018} & Attn+LSTM & C & Attn & P \\
         \textit{STGCN}~\cite{stgcn_2018} & TCN & C & GNN & P  \\
         \textit{GraphWaveNet}~\cite{graphwavenet_2019} & TCN & C & GNN & P+A  \\
         \textit{ASTGCN}~\cite{ASTGCN_2019} & Attn & C-D-W & GNN+Attn & P  \\
         \textit{ST-MGCN}~\cite{geng2019spatiotemporal} & RNN & C & GNN & P  \\
         \textit{GMAN}~\cite{GMAN_2020} & Attn & C & Attn & P \\
         \textit{STSGCN}~\cite{STSGCN_2020} & Attn & C & GNN+Attn & P \\
         \textit{AGCRN}~\cite{AGCRN_2020} & RNN & C & GNN & A\\
         \textit{MTGNN}~\cite{wu2020connecting} & TCN & C & GNN & A\\
         \textit{STMeta}~\cite{STMeta} & RNN & C & GNN & P\\
        \bottomrule
    \end{tabular}
    \label{table: support_models}
    \vspace{-1.5em}
\end{table}

Currently, UCTB has integrated 17 prediction models (Table~\ref{table: support_models}), including 4 statistical learning models (\textit{ARIMA}~\cite{1995Short}, \textit{HM}, \textit{XGBoost}~\cite{xgboost_2016}, etc.) and 13 deep learning models (\textit{ST-ResNet}~\cite{zhang2017deep}, \textit{DCRNN}~\cite{li2018dcrnn_traffic}, \textit {ST-MGCN}~\cite {geng2019spatiotemporal}, \textit {STMeta}~\cite {STMeta}, etc.).
These models have been encapsulated into model classes with internal implementations of the training function (`fit' methods in every model class) and prediction function (`predict' methods in every model class). We also summarize the spatiotemporal domain knowledge utilized by these models in Table~\ref {table: support_models}.

In the statistical learning methods, the historical flow is used as input for the widely-used STP model, \textit {ARIMA}. Historical mean (\textit {HM}) generates predictions by averaging past flows. It considers not only recent data but also same-time flows from previous days and weeks. Both \textit{GBRT} and \textit{XGBoost} utilize features similar to \textit{HM} that reflect various types of temporal knowledge such as closeness, daily periodicity, and weekly periodicity.  
In deep learning methods, \textit{DCRNN} utilizes diffusion convolution and RNN to capture spatial and temporal dependencies. \textit{DCRNN} constructs a distance graph reflecting spatial proximity correlation while \textit{ST-MGCN} builds multiple graphs to capture various kinds of spatial dependencies. Finally, \textit{STMeta} is a spatiotemporal knowledge fusion framework focused on leveraging spatiotemporal knowledge and benefiting from developments in spatial or temporal modeling techniques.
For details on other methods, readers can refer to the UCTB documentation.\footnote{\url{https://uctb.github.io/UCTB/md_file/predictive_tool.html#currently-supported-models}} 

\begin{table}[htbp]
\centering
    \caption{Reusable Layers (T: Temporal, S: Spatial).}
    \begin{tabular}{lc}
        \toprule
         \textbf{Interface} & \textbf{Description} \\ 
         \midrule
         \textit{DCGRU}~\cite{li2018dcrnn_traffic} & Diffusion Convolution (S) + GRU (T) \\
         \textit{GCLSTM}~\cite{chai_multi_graph_2018} & ChebNet (S) + LSTM (T) \\
         \textit{GCL}~\cite{chebnet_2016} & ChebNet (S) \\
         \textit{GAT}~\cite{velickovic2018graph} & Graph Attention Layers (S) \\
        \bottomrule
    \end{tabular}
    \vspace{-0.5em}
    \label{table: modeling_techniques}
\end{table}

Table~\ref{table: modeling_techniques} displays the reusable layers implemented in UCTB. Two spatiotemporal modeling units, \textit{DCGRU} \cite{li2018dcrnn_traffic} and \textit{GCLSTM}\cite{chai_multi_graph_2018}, are employed to capture both temporal and spatial features. They substitute matrix product operations inside traditional RNNs with graph convolution operations. ChebNet \cite{chebnet_2016} implements graph convolutional operations grounded in the Chebyshev polynomial. It leverages the Laplacian matrix to discern spatial interdependencies.  Graph Attention Layer (GAL) \cite{velickovic2018graph} introduces an attention mechanism for graph learning.

\begin{table}[t]
\centering
    \caption{Training and Evaluation Interface.}
    \resizebox{0.475\textwidth}{!}{
    \begin{tabular}{ccccccc}
        \toprule
         \textbf{Module} & \textbf{Interface}  & \textbf{Description}\\ 
         \midrule
        \multirow{2}{*}{\textit{Training}} & 
        \textit{MiniBatchFeedDict} & Mini-batch training mechanism\\
         \cmidrule{2-3}
          & \textit{EarlyStopping} & Early stopping mechanism \\
          \midrule
         \textit{Evaluation} & \textit{RMSE}/\textit{MAE}/\textit{MAPE}/\textit{SMAPE} & Evaluation metrics \\
        \bottomrule
    \end{tabular}}
    \vspace{-2em}
    \label{table: training_evaluating}
\end{table}

\renewcommand{\arraystretch}{1.3}

\subsection{Training and Evaluation Interface}
Table~\ref{table: training_evaluating} outlines the training and evaluation interfaces in UCTB. When applying DL models to large training datasets, the full dataset cannot be read into memory simultaneously. UCTB thus employs mini-batch data to update gradients incrementally over multiple epochs. 
The \textit{MiniBatchFeedDict} interface implements this functionality by continuously invoking the internal \textit{get\_batch} method to generate batch data for training.

UCTB implements two early stopping mechanisms: the naive method and the \textit{t-test} approach. The naive method permits several steps without achieving a lower validation set error. In contrast, the \textit{t-test} approach partitions recent validation set errors into two independent samples, each comprising $n$ samples. An independent samples t-test is then conducted. 
The null hypothesis is that the means of both samples are equivalent. If the p-value of the hypothesis test falls below a threshold (typically 0.1 or 0.05), this hypothesis is rejected, indicating that the model convergence criteria have not yet been met. Otherwise, model convergence is achieved, and training ceases.  

In addition, UCTB integrates four commonly used evaluation metrics in STP prediction including RMSE, MAE, MAPE, and SMAPE (Symmetric Mean Absolute Percentage Error). SMAPE is a symmetric evaluation metric that gives equal weights to overestimate and underestimate.

\section{Experiment and Use Case}
Utilizing UCTB, we implement comprehensive experiments evaluating the effectiveness of our constructed services based on our proposed workflow and introduce a use case for building STP service. 
All experimental codes and tutorials have been made openly available\footnote{https://uctb.github.io/UCTB/md\_file/all\_results.html}.  

\subsection{Experiment}

\subsubsection{Experiment Datasets}
To help researchers better evaluate their STP service, we build a set of STP benchmark datasets across three scenarios shown in \ref{table: benchmark_datasets}. 

\begin{table}[htbp]
\centering
    \caption{Benchmark Dataset Information. (NYC for New York City; CHI for Chicago; DC for Washington DC; BJ for Beijing; XA for Xian; CD for Chengdu; BAY for San Francisco; LA for Los Angeles; SH for Shanghai; CQ for Chongqing; MEL for Melbourne)}
    \begin{tabular}{ccccc}
        \toprule
         \textbf{Scenario} & \textbf{City} & \textbf{Time Span} & \textbf{Interval} & \textbf{\# Units}\\ 
         \midrule
 \textit{Bikesharing}& NYC &2013.07-2017.09&5\&60 mins&820 \\
 \textit{Bikesharing} & CHI & 2013.07-2017.09&5\&60 mins&585 \\
 \textit{Bikesharing} & DC &2013.07-2017.09&5\&60 mins&532   \\
 \textit{EV} & BJ & 2018.03-2018.08&60 mins&629  \\
 \textit{Ridesharing} & XA & 2016.10-2016.11&5 mins&256 \\
 \textit{Ridesharing} & CD & 2016.10-2016.11&5 mins&256 \\
 \textit{Speed} & LA &2012.03-2012.06&5 mins&207   \\
 \textit{Speed} & BAY &2017.01-2017.07&5 mins&325\\
 \textit{Metro} & SH & 2016.07-2016.09&60 mins&288  \\
 \textit{Metro} & CQ & 2016.08-2017.07&60 mins&113  \\
 \textit{Metro} & NYC & 2022.02-2023.12&60 mins&426  \\
 \textit{Bus} & NYC &2022.02-2024.01&60 mins&226  \\
 \textit{Pedestrian} & MEL & 2021.01-2022.11&60 mins&55 \\
 \textit{Taxi} & CHI & 2013.01-2018.01&15 mins&121 \\
 \textit{Taxi} & NYC & 2009.01-2023.06&5 mins&263 \\
        \bottomrule
    \end{tabular}
    \label{table: benchmark_datasets}
    \vspace{-2.5em}
\end{table}

We also gather external factors like weather, holidays, and points of interest (POIs) for these datasets. The original crowd mobility records are converted into UCTB data format with intervals of 15, 30, 60 minutes. Due to space limits, we only report results for three scenarios i.e., Bikesharing, Metro, EV (Bikesharing for bikesharing demand, Metro for metro flow, EV for electric vehicle charging station usage). Additional results can be found in the https://uctb.github.io/UCTB/md\_file/all\_results.html. \label{sec: results}

\subsubsection{Experiment Settings} We summarize experiment settings into three parts, Implementation Details, Train/Validation/Test Split and Hardware and Software.

\textbf{Implementation Details}. Leveraging the capabilities of UCTB, we include temporal knowledge like closeness, daily periodicity and weekly periodicity, alongside spatial knowledge including proximity and functionality. Besides, for our constructed service, \textit{STMeta}, we rapidly define the model with reusable spatiotemporal modeling units and aggregation layers through UCTB. To achieve a fair comparison, we construct every service following our proposed workflow with the same data loader and knowledge management module (if service required) in UCTB to process input. 

\textbf{Train/Validation/Test Split}. We choose the last 10\% duration in each dataset as test data, the 10\% data before the test for validation.

\textbf{Hardware and Software}. Our benchmark platform is a server with an AMD Ryzen 3900x 2.2GHz CPU, a GeForce RTX 2080 Ti GPU, and 64GB RAM. We use Python 3.6 and Tensorflow 1.13.1. More software environment details are in the document\footnote{https://uctb.github.io/UCTB/md\_file/installation.html}.

\subsubsection{Evaluation Metrics}
We employ the RMSE as a metric to report service performance. Additionally, we calculate two aggregation metrics~\cite{STMeta} to indicate the performance across different scenarios. $x \in \mathcal X$ ($\mathcal X$ is the set of all approaches) among all the datasets $\mathcal D$:
\begin{equation}
	\textit{NRMSE}_x  = agg_{d\in \mathcal D}(\frac{RMSE_{x, d}}{min_{x'\in \mathcal X}(RMSE_{x',d})})
\end{equation}
where $RMSE_{x,d}$ is the RMSE of approach $x$ in dataset $d$. We aggregate the normalized RMSE to form the final metric, where normalization is performed relative to the smallest RMSE value within each dataset $d$. When $agg$ takes $avg$ as the aggregation function, the metric means calculating the arithmetic mean of the normalized RMSE values, namely \textit{AvgNRMSE}. When $agg$ takes $max$ as the aggregation function, the metric calculates the maximum of the normalized RMSE values, namely \textit{WstNRMSE}. If a method exhibits generalizability, its \textit{AvgNRMSE} and \textit{WstNRMSE} should approach 1.

\subsubsection{Evaluation}
\begin{figure}[htbp]
  \centering
  \includegraphics[width=.83\linewidth]{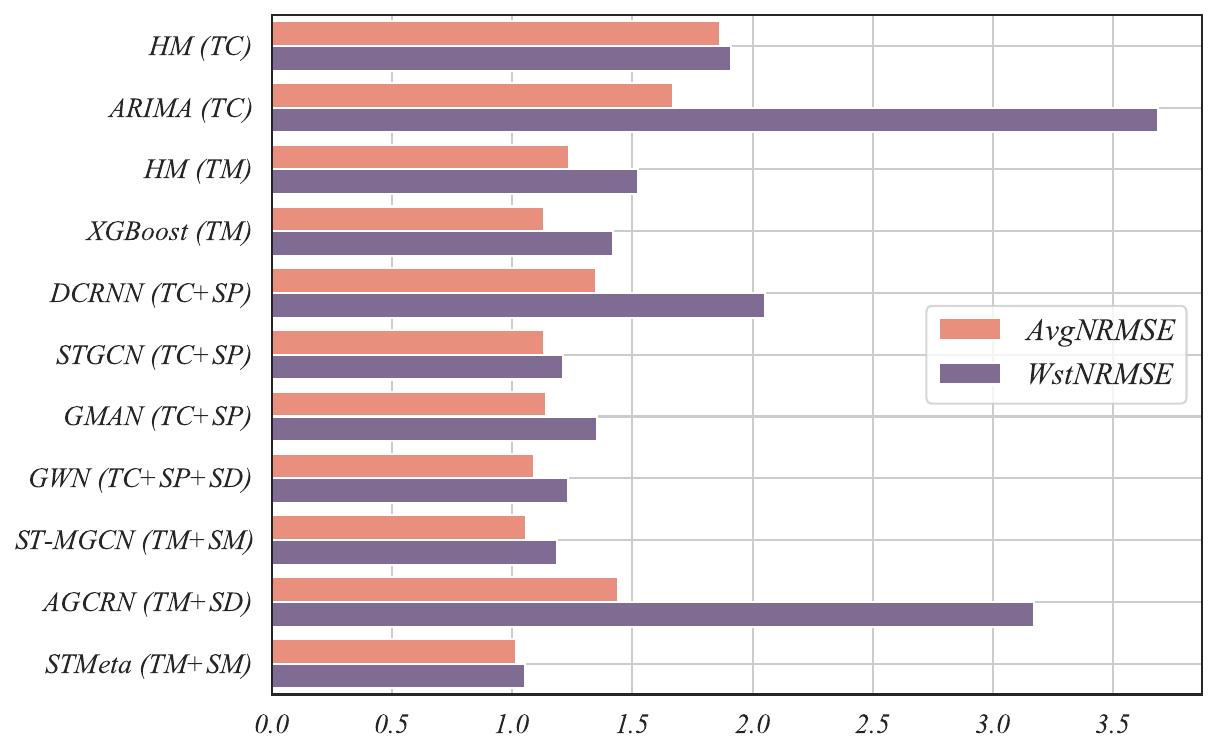}
  \vspace{-.5em}
  \caption{Overall Performance across 3 Scenarios (Bikesharing, Metro, EV). (TC: Temporal Closeness; TM: Multi-Temporal Factors; SP: Spatial Proximity; SM: Multi-Spatial Factors; SD: Data-driven Spatial Knowledge Extraction.)}
  \label{fig:benchmark}
  \vspace{-1em}
\end{figure}

In Figure~\ref{fig:benchmark}, we present the evaluation results for 30-minute spatiotemporal prediction. We also evaluate the services on STP tasks with the 60-minute interval and the 15-minute interval. Due to the limit of pages, full results are on https://uctb.github.io/UCTB/md\_file/all\_results.html. The \textit{avgNRMSE} and \textit{WstNRMSE} metrics are calculated to assess overall performance across diverse scenarios. For each service, the temporal and/or spatial factors considered are specified. 
We have three key observations from the results. First, taking into account multiple kinds of temporal knowledge as well as spatial knowledge, \textit{STMeta}) achieves the best performance, highlighting the importance of integrating and managing domain knowledge.
Second, some services have experienced significant improvements (more than 20\%) with integration of multiple knowledge through UCTB compared to the original services e.g. \textit{HM (TC)} vs \textit{HM (TM)}, which can be even comparable to some deep models using single knowledge (\textit{DCRNN}, \textit{GMAN}).  Third, models with data-driven spatial knowledge extraction (\textit{GraphWaveNet} and \textit{AGCRN}) also demonstrate relatively superior performance with second-best \textit{AvgNRMSE} of \textit{GraphWaveNet} model.

\newcommand{\fourfigcol}{-2mm}
\begin{figure*}[htbp]
\centering
\subfigure[Temporal Knowledge]{
\includegraphics[width=0.32\linewidth]{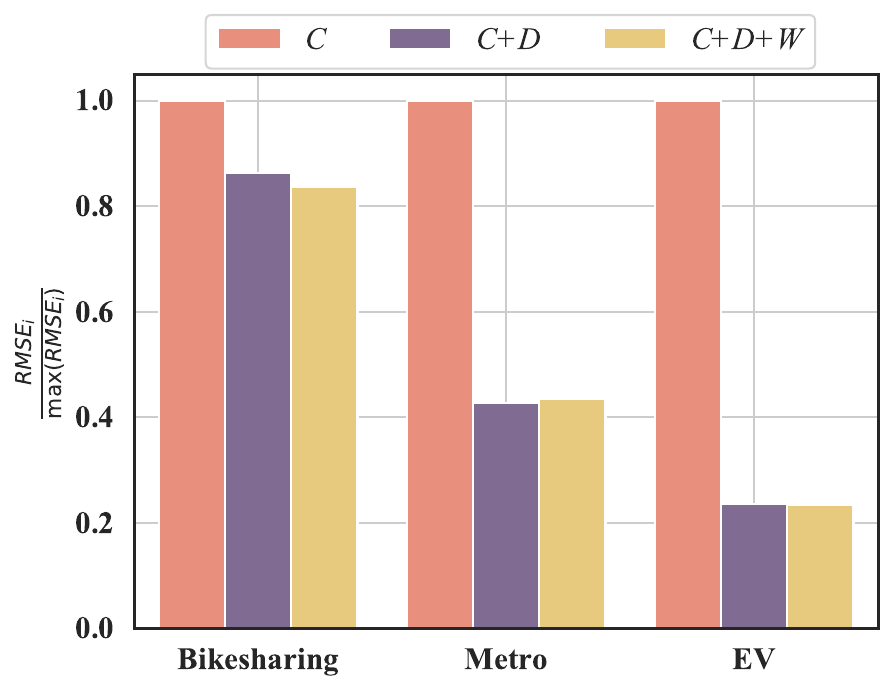}
\label{fig: temporal_k}
}\hspace{\fourfigcol}
\subfigure[Spatial Knowledge]{
\includegraphics[width=0.32\linewidth]{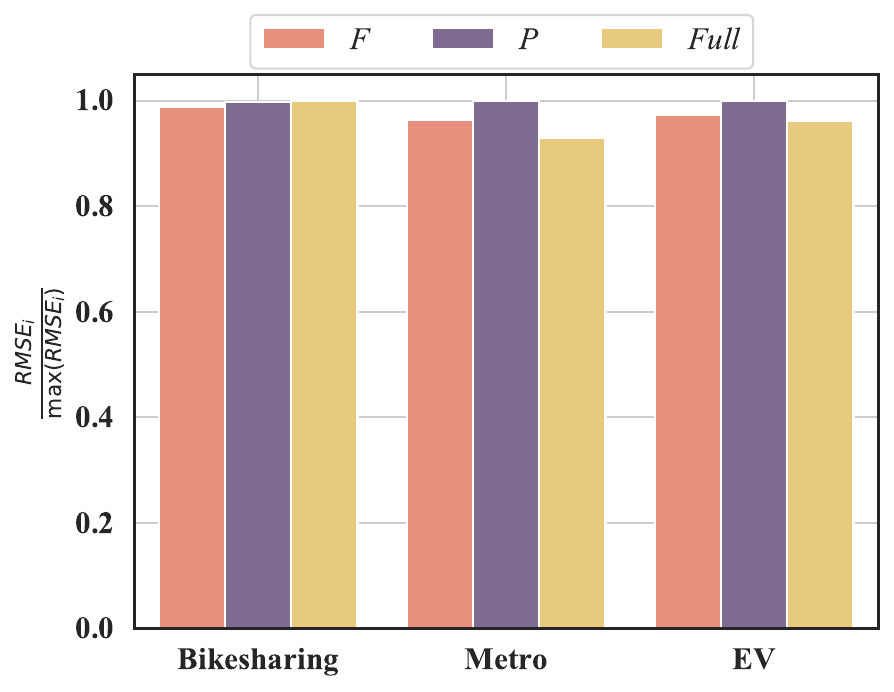}
\label{fig: spatial_k}}
\hspace{\fourfigcol}
\subfigure[External Factors]{
\includegraphics[width=0.32\linewidth]{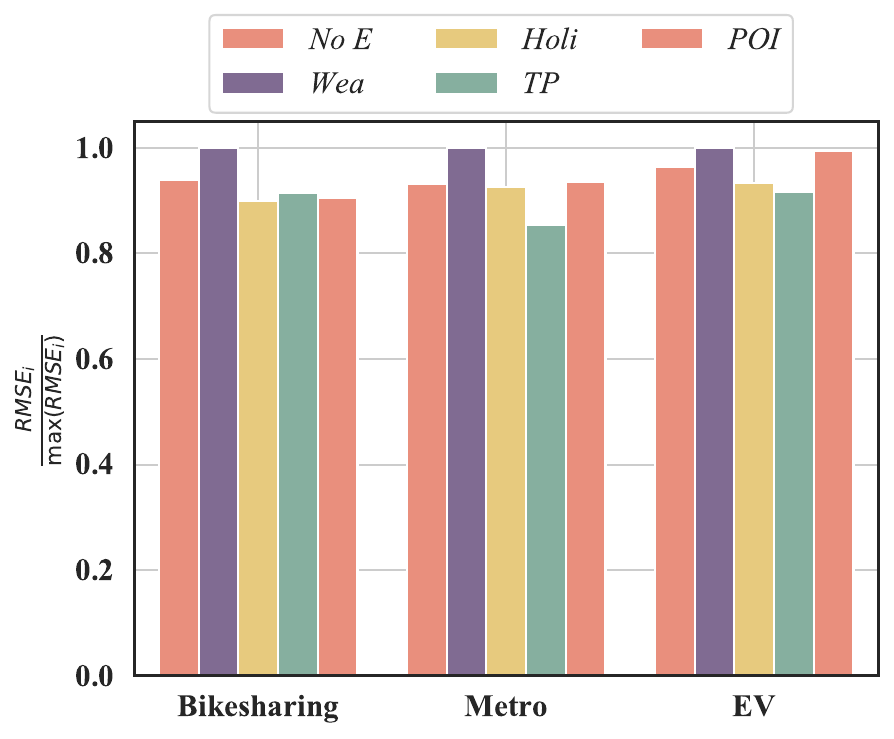}
\label{fig: external}
}
\vspace{-0.8em}
\caption{Knowledge Management across Different Scenarios.}
\label{knowledge_selection}
\vspace{-1.7em}
\end{figure*}

\subsubsection{Analysis of Knowledge Management} To show how different selections of knowledge (i.e., temporal, spatial, external knowledge) impact the prediction, we conduct the following comparison in the following three aspects.

For temporal knowledge, we use the knowledge management module to create three combinations (\textit{C}: Closeness, \textit{C+D}: Closeness and Daily Periodicity, \textit{C+D+W}: Closeness, Daily Periodicity, Weekly Periodicity). From Figure~\ref{fig: temporal_k}, we observe that adding periodicity can always enhance the prediction accuracy. However, the extent of performance enhancement from incorporating periodicity knowledge varies depending on the scenarios.

For spatial knowledge, we remove some spatial factors (F for Functionality and P for Proximity) from \textit{STMeta-DCG-GAL} and then compare it to the original \textit{STMeta-DCG-GAL} (with full spatial knowledge). From Figure~\ref{fig: spatial_k}, we observe that among single spatial graphs considered, the functionality graph is often better than the other one. However, it is not the same as temporal knowledge that more knowledge considered brings about more performance promotion. In the scenario of Bikesharing, full use of spatial knowledge even causes the worst performance among the spatial knowledge combinations.

For external factors, we incorporate 5 categories of external factors (\textit{No E}: Without External Factors, \textit{Wea}: Weather, \textit{Holi}: Holiday, \textit{TP}: Temporal Position, \textit{POI}: Point of Interest) for STP. From Figure~\ref{fig: external} we observe that not all external factors can make better predictions. Weather performs worse than without external factors across all three scenarios within the 30-minute-time-interval STP tasks, while holiday and temporal position bring the greatest promotion.

\subsubsection{Analysis of Reusable Layers} 
\begin{figure}[h]
  \centering
  \includegraphics[width=.7\linewidth]{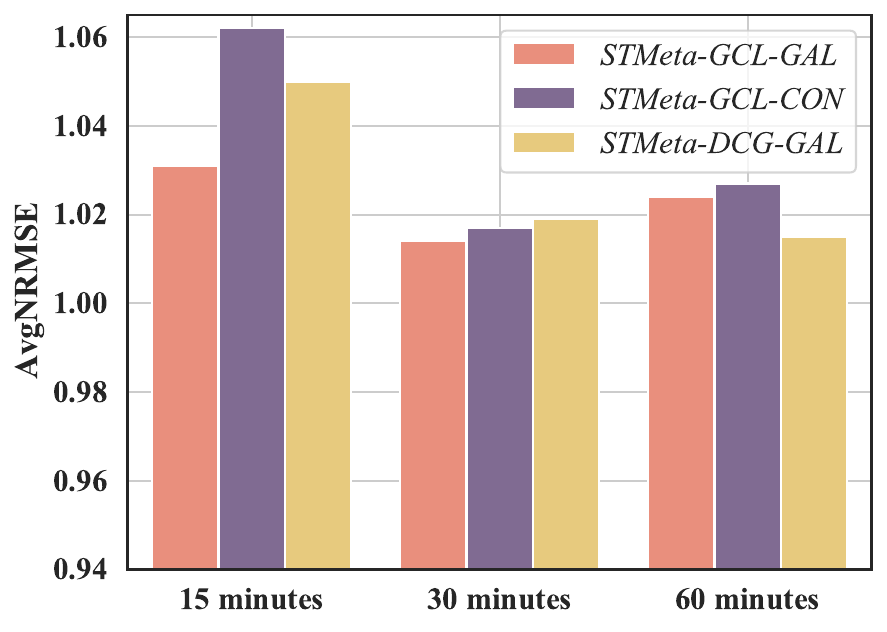}
  \caption{\textit{AvgNRMSE} Comparison of Different Reusable Layers across 3 Scenarios (Bikesharing, Metro, EV).}
  \label{fig: model_generalize}
  \vspace{-2em}
\end{figure}

To analyze the influence of modeling techniques, we test the performance of three model instances (\textit{STMeta-GCL-GAL}, \textit{STMeta-GCL-CON}, \textit{STMeta-DCG-GAL}) on three datasets at three temporal granularities shown in Figure~\ref{fig: model_generalize}. Based on the results of various prediction granularities, \textit{STMeta-GCL-GAL} has the best generalization performance and is more likely to perform better in STP tasks with different temporal granularities.

\subsubsection{Guidelines}\label{sec: guide}
The experiment analysis reveals key insights into prediction accuracy based on different knowledge types and modeling techniques. Incorporating periodicity consistently improves predictions, although the degree of enhancement varies across scenarios. Spatial knowledge, particularly functionality, influences prediction accuracy, with full spatial knowledge sometimes leading to suboptimal results, notably in Bikesharing scenarios. External factors like holidays and temporal position significantly enhance predictions, while weather often hinders accuracy. Among modeling techniques, \textit{STMeta-GCL-GAL} shows superior generalization performance, indicating its potential for better outcomes in new scenarios.

\subsection{Example Use Case}
To illustrate the ability of our toolbox, we apply it to the case of bus ridership prediction in NYC. The Metropolitan Transportation Authority (MTA) has collected the number of passengers who board buses (i.e. bus ridership) since February 2022\footnote{https://new.mta.info/about}. The raw data\footnote{https://data.ny.gov/Transportation/MTA-Bus-Hourly-Ridership-Beginning-February-2022/kv7t-n8in/about\_data} offers bus ridership estimates on an hourly basis for every individual bus route in New York City. Therefore, our prediction task will be carried out at 60-minute intervals for individual bus routes.

\subsubsection{Data Conversion}
\begin{figure}[htbp]
  \centering
  \includegraphics[width=0.95\linewidth]{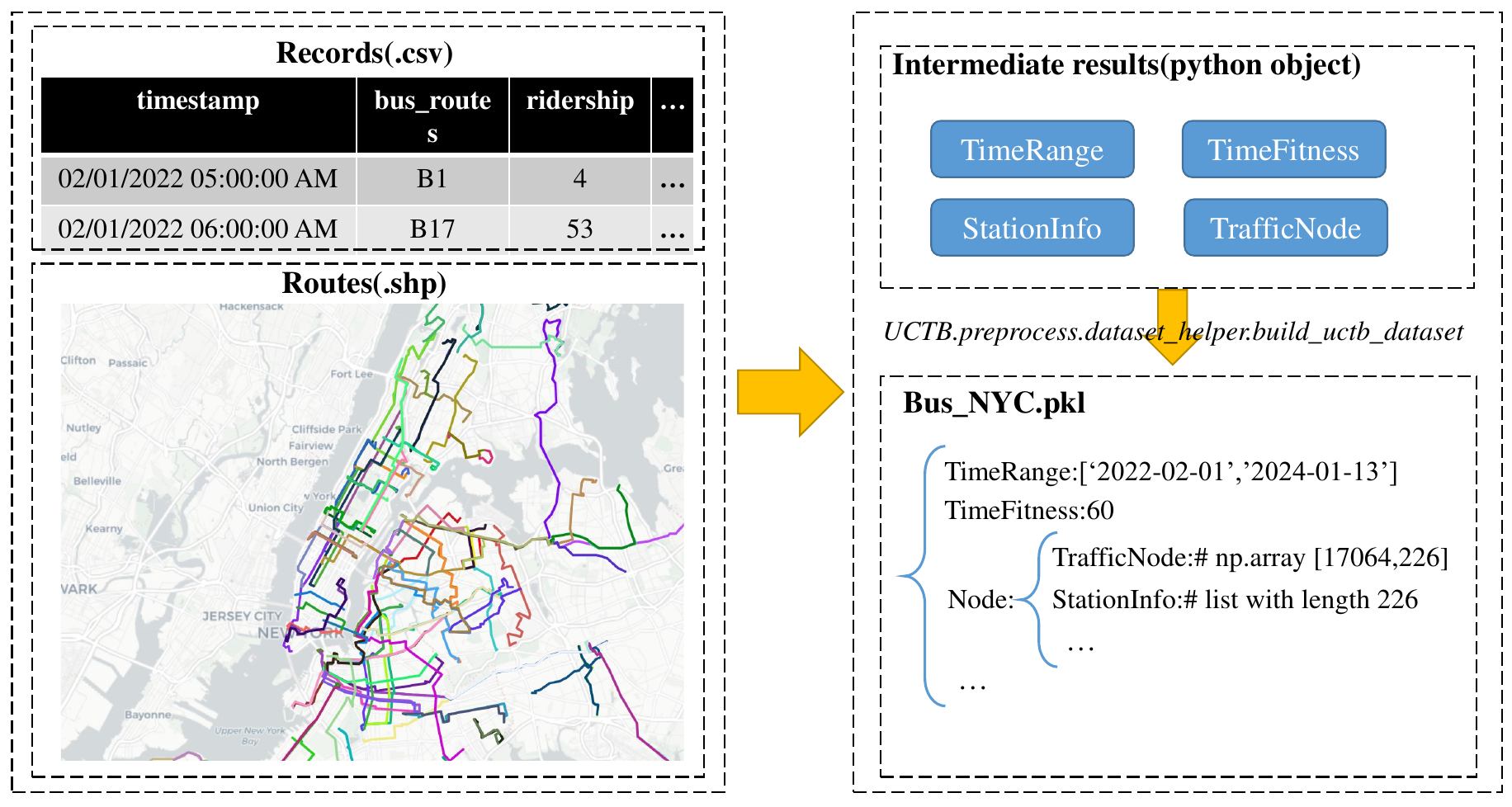}
  \vspace{-0.5em}
  \caption{Example Data Conversion: Bus NYC.}
  \label{fig: data_conversion}
  \vspace{-0.5em}
\end{figure}

According to existing literature~\cite{zhang2017deep, li2018dcrnn_traffic, STMeta}, we collect bus ridership data from 2022.02.01 to 2023.02.01 (i.e. \textbf{TimeRange}) for a year. Adhering to the granularity specifications of our task,  we set \textbf{TimeFitness} to 60 minutes. In order to obtain \textbf{TrafficNode}, we convert the collected data into a two-dimensional matrix structure, whose cell encapsulates the ridership data of a specific spatial unit (i.e. bus route) during a particular hour within the aforementioned one-year range. Besides, we download geospatial information of NYC bus routes from the website\footnote{https://archive.nyu.edu/handle/2451/60058}. After filtering out bus routes that appear in both bus ridership data and geospatial information, we calculate the geographical center of bus routes as the coordinates of corresponding spatial units to form \textbf{StationInfo}. Having prepared the intermediate results shown in Figure~\ref{fig: data_conversion}, we obtain data in UCTB data format after calling the \textit{UCTB.preprocess.dataset\_helper.build\_uctb\_dataset} interface. 

\begin{figure}[h]
  \centering
  \includegraphics[width=.95\linewidth]{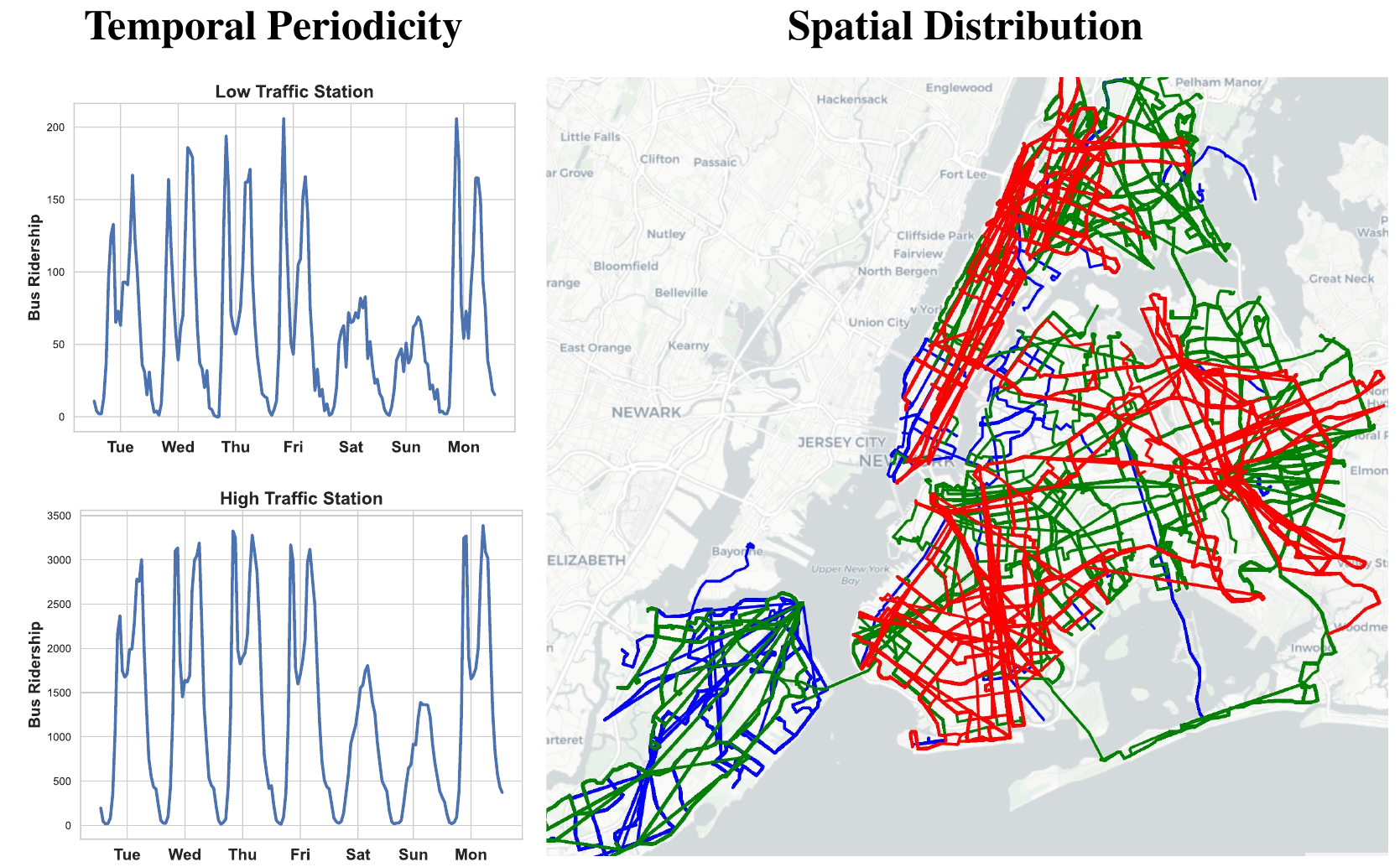}
  \caption{Temporal and Spatial View of Bus Ridership in NYC.}
  \label{fig:dataset_overview}
  \vspace{-0.5em}
\end{figure}

To check the correctness of data conversion and inspire the next procedures, we visualize the data from two perspectives: temporal periodicity and the spatial distribution of spatial units based on ridership volumes (red for high; green for middle; blue for low) as shown in Figure \ref{fig:dataset_overview}. The notable impact of daily periodicity and the days of the week is obvious. Additionally, the spatial distribution aligns with our common sense, as high-traffic units are predominantly situated in downtown, while low-traffic units are mainly found in suburban areas.

\subsubsection{Region Generation}
\begin{figure}[h]
  \centering
  \includegraphics[width=0.95\linewidth]{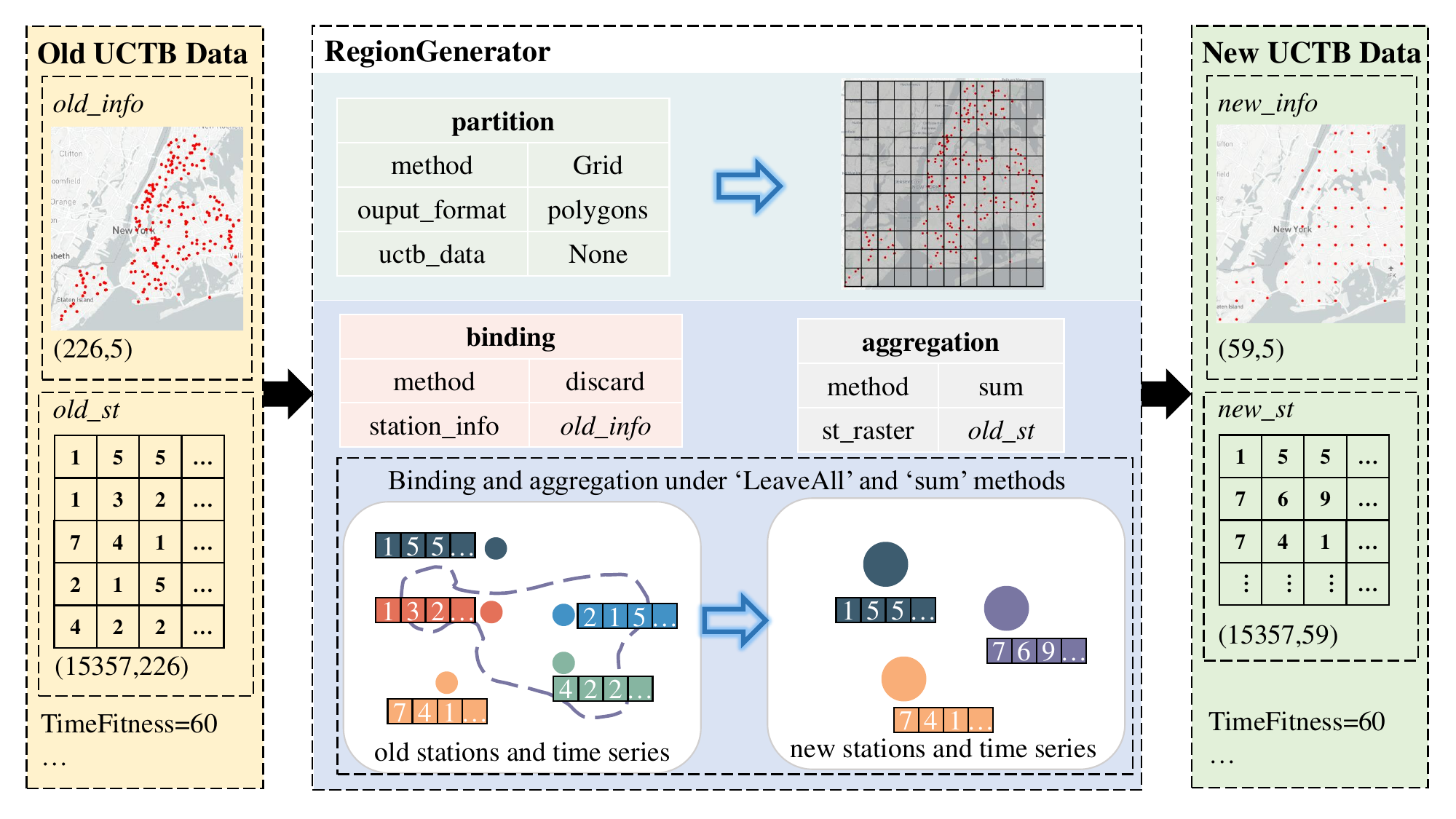}
  \vspace{-0.5em}
  \caption{Instance of Region Generation.}
  \label{fig: Ins_of_RG}
  \vspace{-1.3em}
\end{figure}

The whole process is shown in Figure~\ref{fig: Ins_of_RG}. We set \textit{method} in partitioner to ``Grid'' which does not need spatiotemporal data to assist in partition. For the binding part, we calculate the positional relationship between units and regions and bind each region with units in it. If there are any regions containing no units or any units that do not belong to any regions, we will discard them. Finally, in the aggregation part, we select \textit{sum} as the aggregation method. After region generation, the number of spatial units decreases from 226 to 59.

\subsubsection{Knowledge Management} 
\begin{figure}[h]
  \centering
  \includegraphics[width=0.8\linewidth]{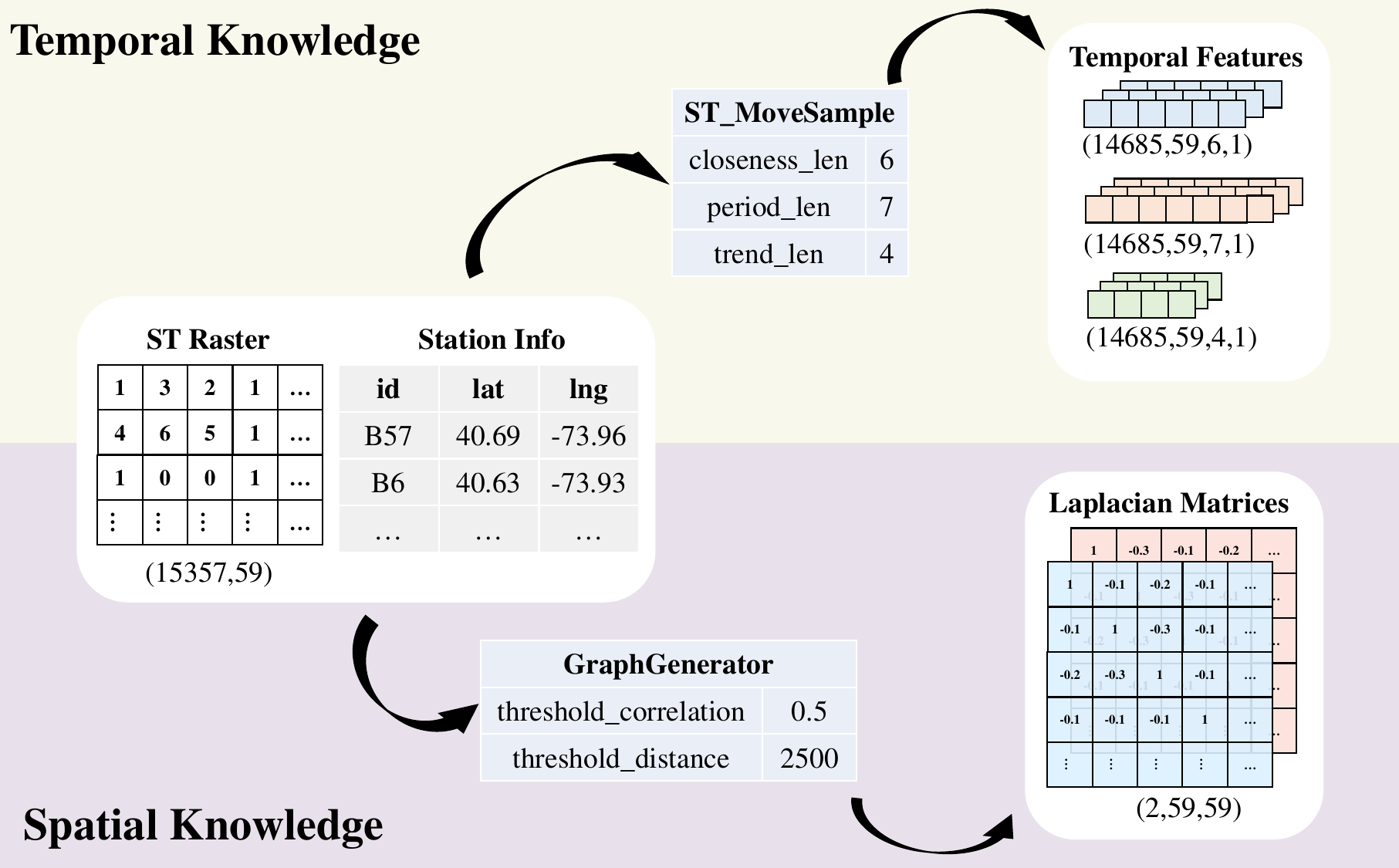}
  \caption{Instance of Knowledge Management.}
  \label{fig: Instance of Knowledge Management}
  \vspace{-0.5em}
\end{figure}

Based on guidelines from Section \ref{sec: guide}, we select closeness, daily periodicity, and weekly periodicity for temporal knowledge (upper part of Figure~\ref{fig: Instance of Knowledge Management}), functionality and proximity graphs for spatial knowledge (lower part of Figure~\ref{fig: Instance of Knowledge Management}). Additionally, we incorporate \textit{temporal position} (i.e., day of week) as the external factor to boost prediction due to insights from Figure~\ref{fig:dataset_overview}. After that, we call \textit{ST\_MoveSample} and \textit{GraphGenerator}) to generate knowledge-driven input with configurations of \textit{closeness\_len}=6, \textit{period\_len}=7, \textit{trend\_len}=4, \textit{Graph}=``Correlation-Distance'', \textit{threshold\_correlation}=0.5, \textit{threshold\_distance}=2500.

\subsubsection{Model Definition}
\begin{figure}[h]
  \centering
  \includegraphics[width=0.95\linewidth]{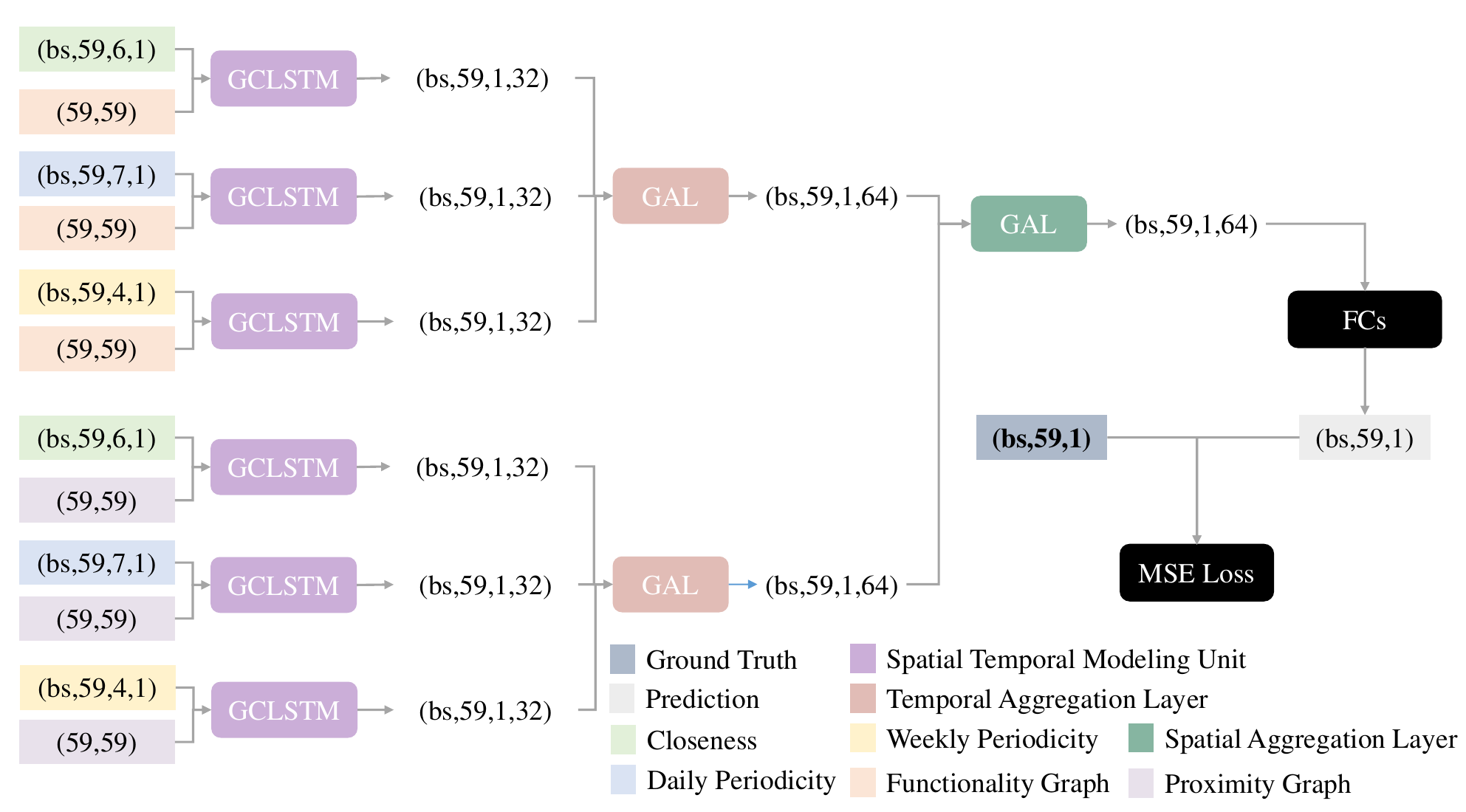}
  \caption{Instance of \textit{STMeta} (``bs'' for \textit{batch\_size}). }
  \label{fig: real_paradigm}
  \vspace{-0.5em}
\end{figure}

Based on the guidelines in Section \ref{sec: guide}, we choose \textit{GCLSTM} as the spatiotemporal modeling unit, \textit{GAL} as the temporal aggregation unit, \textit{GAL} as the spatial aggregation unit. The dimension of \textit{GCLSTM}'s hidden states is set to 32 and the dimension of \textit{GAL}'s hidden states is 64. After several fully connected layers, we obtain the final predictions and calculate MSE loss with ground truths. The instance of \textit{STMeta} model we construct is shown in Figure~\ref{fig: real_paradigm} (The ``bs'' means \textit{batch\_size}).

\begin{figure*}[htbp]
\centering
\subfigure[Spatial Unit Visualization]{
\includegraphics[width=0.377\linewidth]{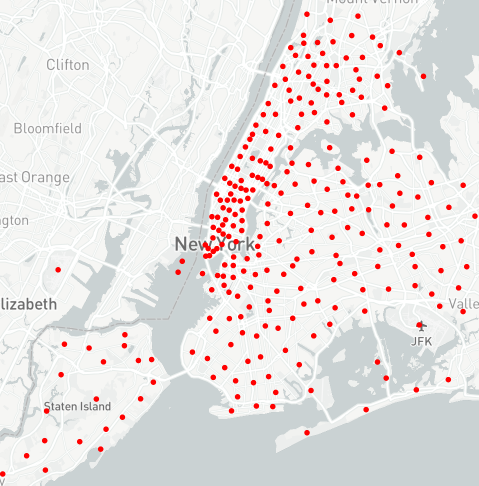}
\label{fig: station_vis}
}\hspace{\fourfigcol}
\subfigure[Prediction Results Visualization]{
\includegraphics[width=0.58\linewidth]{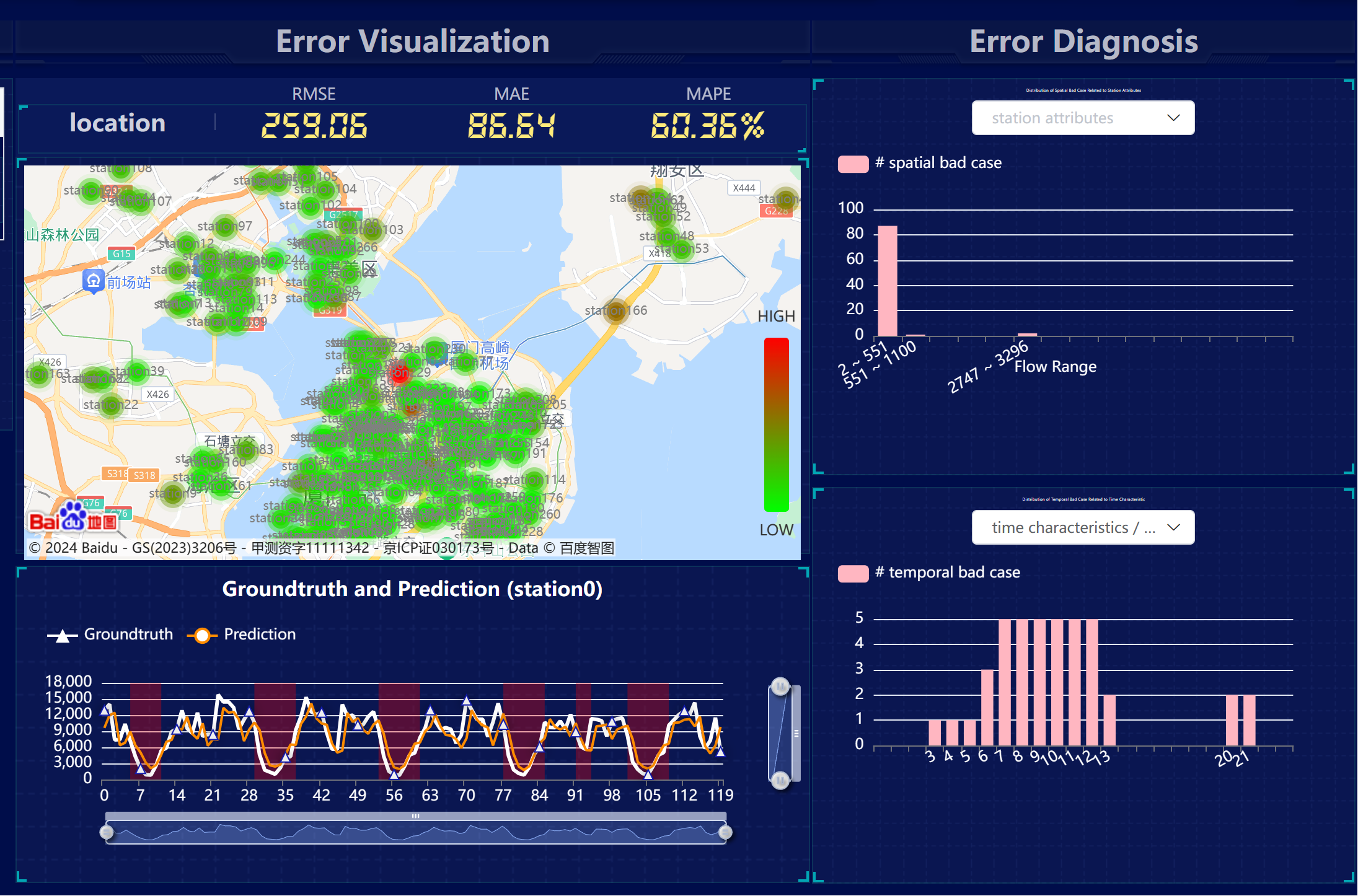}
\label{fig: error_vis}
}
\vspace{-0.7em}
\caption{Visualization Tools in UCTB.}
\vspace{-1.5em}
\end{figure*}

\subsubsection{Training and Evaluation} 
\begin{figure}[h]
  \centering
  \includegraphics[width=0.95\linewidth]{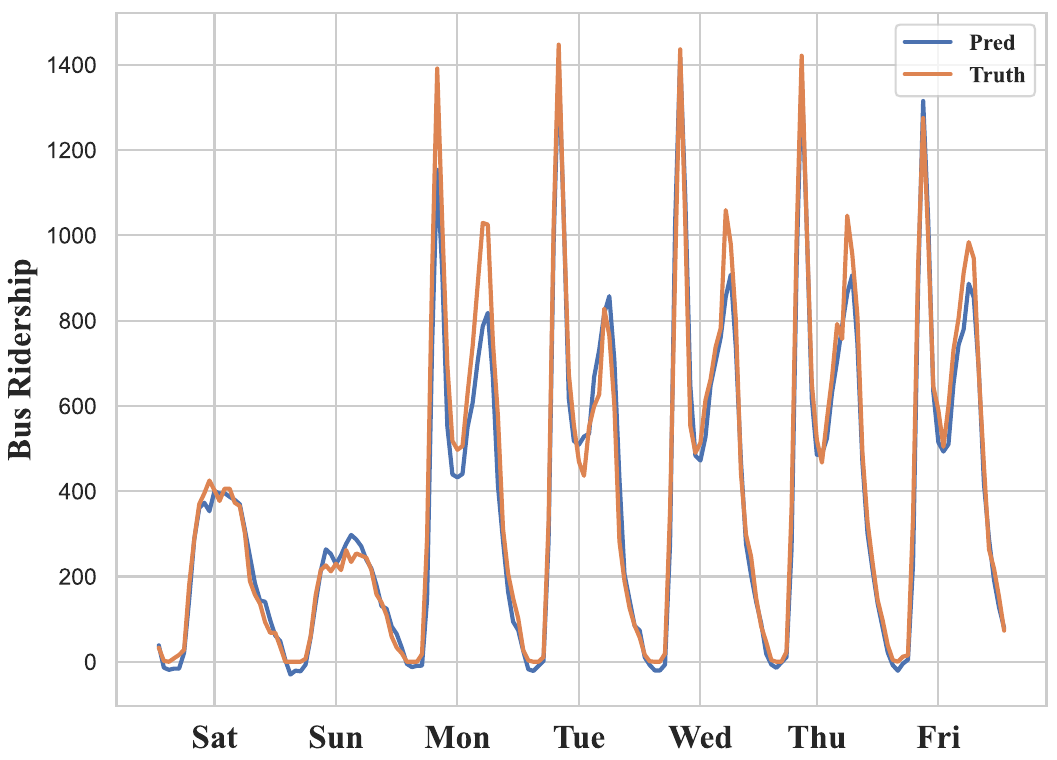}
  \caption{Visualization of Prediction Results.}
  \vspace{-1em}
  \label{fig: pred_vs_truth}

\end{figure}

We combine the previous code snippets to form an experimental script. Please refer to the detailed parameter settings in the UCTB repository. However, the parameters \textit{learning\_rate}, \textit{early\_stop\_patience}, \textit{early\_stop\_length} are sensitive to scenarios and highly correlated with the convergence of STP models. We choose the proper ones from our debug experience. In the evaluation stage, we select RMSE, MAE, MAPE, and SMAPE as metrics to assess the performance of our STP service. We set the proper threshold for these evaluation metrics through an analysis of the distribution of ridership data. The final evaluation result is RMSE=80.63, MAE=47.03, MAPE=0.194, SMAPE=0.237. We visualize the prediction result of a single unit within the range of a week as shown in Figure~\ref{fig: pred_vs_truth}

\section{Visualization}
To enhance users' better understanding of datasets and experimental results, we have designed two types of visualization interfaces. The first type is a spatial unit visualization interface that shows the spatial locations of each unit. We have integrated this function into \textit{NodeTrafficLoader} and \textit{GridTrafficLoader}. By calling their \textit{st\_map} method, users could get the units visualization as shown in Figure~\ref{fig: station_vis}. 
Additionally, UCTB offers interactive interfaces for visualizing experimental results from various models and units. These interfaces were implemented using JavaScript, with the demo UI page displayed in Figure~\ref{fig: error_vis}. More details are in our repository\footnote{https://github.com/uctb/visualization-tool} and our preliminary demo is deployed at http://39.107.116.221/.

\section{Conclusion}
We propose a novel workflow for building STP services, develop and release an open-source toolbox based on the proposed workflow, UCTB, to enable to accelerate developing STP services for new scenarios. 
Through UCTB, we aim to improve spatial units partition, facilitate comprehensive utilization of domain knowledge, foster sophisticated model development,  and promote reproducible research in the spatiotemporal forecasting domain. While our study makes contributions to STP service construction, we don't carefully evaluate the ease of use and extensibility of UCTB in a qualitative and quantitative manner. Additionally, the current workflow lacks automated tools to assist with the data conversion process. In summary, by providing an array of predictive models and tools for knowledge incorporation in an open and well-documented framework, we look to lower barriers to progress and catalyze innovative solutions to predictive challenges. We view UCTB as an evolving community resource and welcome user comments and contributions to help advance, expand, and strengthen the toolbox.


\bibliographystyle{IEEEtran}
\bibliography{ref}

\begin{thebibliography}{10}
\providecommand{\url}[1]{#1}
\csname url@samestyle\endcsname
\providecommand{\newblock}{\relax}
\providecommand{\bibinfo}[2]{#2}
\providecommand{\BIBentrySTDinterwordspacing}{\spaceskip=0pt\relax}
\providecommand{\BIBentryALTinterwordstretchfactor}{4}
\providecommand{\BIBentryALTinterwordspacing}{\spaceskip=\fontdimen2\font plus
\BIBentryALTinterwordstretchfactor\fontdimen3\font minus \fontdimen4\font\relax}
\providecommand{\BIBforeignlanguage}[2]{{%
\expandafter\ifx\csname l@#1\endcsname\relax
\typeout{** WARNING: IEEEtran.bst: No hyphenation pattern has been}%
\typeout{** loaded for the language `#1'. Using the pattern for}%
\typeout{** the default language instead.}%
\else
\language=\csname l@#1\endcsname
\fi
#2}}
\providecommand{\BIBdecl}{\relax}
\BIBdecl

\bibitem{chen2019review}
X.~Chen and R.~Chen, ``A review on traffic prediction methods for intelligent transportation system in smart cities,'' in \emph{2019 12th International Congress on Image and Signal Processing, BioMedical Engineering and Informatics (CISP-BMEI)}.\hskip 1em plus 0.5em minus 0.4em\relax IEEE, 2019, pp. 1--5.

\bibitem{yuan2021survey}
H.~Yuan and G.~Li, ``A survey of traffic prediction: from spatio-temporal data to intelligent transportation,'' \emph{Data Science and Engineering}, vol.~6, pp. 63--85, 2021.

\bibitem{zheng2014urban}
Y.~Zheng, L.~Capra, O.~Wolfson, and H.~Yang, ``Urban computing: concepts, methodologies, and applications,'' \emph{ACM Transactions on Intelligent Systems and Technology (TIST)}, vol.~5, no.~3, pp. 1--55, 2014.

\bibitem{ASTGCN_2019}
S.~Guo, Y.~Lin, N.~Feng, C.~Song, and H.~Wan, ``Attention based spatial-temporal graph convolutional networks for traffic flow forecasting,'' \emph{AAAI}, 2019.

\bibitem{graphwavenet_2019}
Z.~Wu, S.~Pan, G.~Long, J.~Jiang, and C.~Zhang, ``Graph wavenet for deep spatial-temporal graph modeling,'' in \emph{IJCAI}, 2019.

\bibitem{mtgnn}
Z.~Wu, S.~Pan, G.~Long, J.~Jiang, X.~Chang, and C.~Zhang, ``Connecting the dots: Multivariate time series forecasting with graph neural networks,'' in \emph{Proceedings of the 26th ACM SIGKDD international conference on knowledge discovery \& data mining}, 2020, pp. 753--763.

\bibitem{attention_2017}
\BIBentryALTinterwordspacing
A.~Vaswani, N.~Shazeer, N.~Parmar, J.~Uszkoreit, L.~Jones, A.~N. Gomez, L.~u. Kaiser, and I.~Polosukhin, ``Attention is all you need,'' in \emph{Advances in Neural Information Processing Systems}, I.~Guyon, U.~V. Luxburg, S.~Bengio, H.~Wallach, R.~Fergus, S.~Vishwanathan, and R.~Garnett, Eds., vol.~30.\hskip 1em plus 0.5em minus 0.4em\relax Curran Associates, Inc., 2017. [Online]. Available: \url{https://proceedings.neurips.cc/paper_files/paper/2017/file/3f5ee243547dee91fbd053c1c4a845aa-Paper.pdf}
\BIBentrySTDinterwordspacing

\bibitem{jin2022deep}
G.~Jin, Z.~Xi, H.~Sha, Y.~Feng, and J.~Huang, ``Deep multi-view graph-based network for citywide ride-hailing demand prediction,'' \emph{Neurocomputing}, vol. 510, pp. 79--94, 2022.

\bibitem{ji2022stden}
J.~Ji, J.~Wang, Z.~Jiang, J.~Jiang, and H.~Zhang, ``Stden: Towards physics-guided neural networks for traffic flow prediction,'' in \emph{Proceedings of the AAAI Conference on Artificial Intelligence}, vol.~36, no.~4, 2022, pp. 4048--4056.

\bibitem{wang2021gallat}
Y.~Wang, H.~Yin, T.~Chen, C.~Liu, B.~Wang, T.~Wo, and J.~Xu, ``Gallat: A spatiotemporal graph attention network for passenger demand prediction,'' in \emph{2021 IEEE 37th International Conference on Data Engineering (ICDE)}.\hskip 1em plus 0.5em minus 0.4em\relax IEEE, 2021, pp. 2129--2134.

\bibitem{zhang2021traffic}
X.~Zhang, C.~Huang, Y.~Xu, L.~Xia, P.~Dai, L.~Bo, J.~Zhang, and Y.~Zheng, ``Traffic flow forecasting with spatial-temporal graph diffusion network,'' in \emph{Proceedings of the AAAI conference on artificial intelligence}, vol.~35, no.~17, 2021, pp. 15\,008--15\,015.

\bibitem{shao2022decoupled}
Z.~Shao, Z.~Zhang, W.~Wei, F.~Wang, Y.~Xu, X.~Cao, and C.~S. Jensen, ``Decoupled dynamic spatial-temporal graph neural network for traffic forecasting,'' \emph{arXiv preprint arXiv:2206.09112}, 2022.

\bibitem{li2023dynamic}
F.~Li, J.~Feng, H.~Yan, G.~Jin, F.~Yang, F.~Sun, D.~Jin, and Y.~Li, ``Dynamic graph convolutional recurrent network for traffic prediction: Benchmark and solution,'' \emph{ACM Transactions on Knowledge Discovery from Data}, vol.~17, no.~1, pp. 1--21, 2023.

\bibitem{lan2022dstagnn}
S.~Lan, Y.~Ma, W.~Huang, W.~Wang, H.~Yang, and P.~Li, ``Dstagnn: Dynamic spatial-temporal aware graph neural network for traffic flow forecasting,'' in \emph{International conference on machine learning}.\hskip 1em plus 0.5em minus 0.4em\relax PMLR, 2022, pp. 11\,906--11\,917.

\bibitem{fang2021spatial}
Z.~Fang, Q.~Long, G.~Song, and K.~Xie, ``Spatial-temporal graph ode networks for traffic flow forecasting,'' in \emph{Proceedings of the 27th ACM SIGKDD conference on knowledge discovery \& data mining}, 2021, pp. 364--373.

\bibitem{lin2023predicting}
K.-Y. Lin, P.-Y. Liu, P.-K. Wang, C.-L. Hu, and Y.~Cai, ``Predicting road traffic risks with cnn-and-lstm learning over spatio-temporal and multi-feature traffic data,'' in \emph{2023 IEEE International Conference on Software Services Engineering (SSE)}.\hskip 1em plus 0.5em minus 0.4em\relax IEEE, 2023, pp. 305--311.

\bibitem{miao2017towards}
H.~Miao, A.~Li, L.~S. Davis, and A.~Deshpande, ``Towards unified data and lifecycle management for deep learning,'' in \emph{2017 IEEE 33rd International Conference on Data Engineering (ICDE)}.\hskip 1em plus 0.5em minus 0.4em\relax IEEE, 2017, pp. 571--582.

\bibitem{wang2020deep}
S.~Wang, J.~Cao, and S.~Y. Philip, ``Deep learning for spatio-temporal data mining: A survey,'' \emph{IEEE transactions on knowledge and data engineering}, vol.~34, no.~8, pp. 3681--3700, 2020.

\bibitem{baltensperger2022continuous}
J.~Baltensperger, P.~Salza, and H.~C. Gall, ``Continuous deep learning: A workflow to bring models into production,'' \emph{arXiv preprint arXiv:2208.12308}, 2022.

\bibitem{chen2023data}
L.~Chen, J.~Fang, Z.~Yu, Y.~Tong, S.~Cao, and L.~Wang, ``A data-driven region generation framework for spatiotemporal transportation service management,'' in \emph{Proceedings of the 29th ACM SIGKDD Conference on Knowledge Discovery and Data Mining}, 2023, pp. 3842--3854.

\bibitem{jin2022gridtuner}
J.~Jin, P.~Cheng, L.~Chen, X.~Lin, and W.~Zhang, ``Gridtuner: Reinvestigate grid size selection for spatiotemporal prediction models,'' in \emph{2022 IEEE 38th International Conference on Data Engineering (ICDE)}.\hskip 1em plus 0.5em minus 0.4em\relax IEEE, 2022, pp. 1193--1205.

\bibitem{jin2023spatio}
G.~Jin, Y.~Liang, Y.~Fang, Z.~Shao, J.~Huang, J.~Zhang, and Y.~Zheng, ``Spatio-temporal graph neural networks for predictive learning in urban computing: A survey,'' \emph{IEEE Transactions on Knowledge and Data Engineering}, 2023.

\bibitem{zhang2017deep}
J.~Zhang, Y.~Zheng, and D.~Qi, ``Deep spatio-temporal residual networks for citywide crowd flows prediction,'' in \emph{AAAI}, 2017.

\bibitem{deepstn_2019}
Z.~Lin, J.~Feng, Z.~Lu, Y.~Li, and D.~Jin, ``Deepstn+: Context-aware spatial-temporal neural network for crowd flow prediction in metropolis,'' \emph{AAAI}, 2019.

\bibitem{xie2021survey}
X.~Xie, J.~Niu, X.~Liu, Z.~Chen, S.~Tang, and S.~Yu, ``A survey on incorporating domain knowledge into deep learning for medical image analysis,'' \emph{Medical Image Analysis}, vol.~69, p. 101985, 2021.

\bibitem{}
A.~Thusoo, J.~S. Sarma, N.~Jain, Z.~Shao, P.~Chakka, N.~Zhang, S.~Antony, H.~Liu, and R.~Murthy, ``Hive - a petabyte scale data warehouse using hadoop,'' in \emph{2010 IEEE 26th International Conference on Data Engineering (ICDE 2010)}, 2010, pp. 996--1005.

\bibitem{STMeta}
L.~Wang, D.~Chai, X.~Liu, L.~Chen, and K.~Chen, ``Exploring the generalizability of spatio-temporal traffic prediction: Meta-modeling and an analytic framework,'' \emph{IEEE Transactions on Knowledge and Data Engineering}, vol.~35, no.~4, pp. 3870--3884, 2021.

\bibitem{li2018dcrnn_traffic}
Y.~Li, R.~Yu, C.~Shahabi, and Y.~Liu, ``Diffusion convolutional recurrent neural network: Data-driven traffic forecasting,'' in \emph{ICLR '18}, 2018.

\bibitem{STSGCN_2020}
C.~Song, Y.~Lin, S.~Guo, and H.~Wan, ``Spatial-temporal synchronous graph convolutional networks: A new framework for spatial-temporal network data forecasting,'' in \emph{AAAI 2020}, 2020.

\bibitem{stgcn_2018}
B.~Yu, H.~Yin, and Z.~Zhu, ``Spatio-temporal graph convolutional networks: A deep learning framework for traffic forecasting,'' in \emph{IJCAI}, 2018.

\bibitem{yao_similarity_2019}
H.~Yao, X.~Tang, H.~Wei, G.~Zheng, and Z.~Li, ``Revisiting spatial-temporal similarity: A deep learning framework for traffic prediction,'' in \emph{AAAI}, 2019.

\bibitem{yang2014modeling}
D.~Yang, D.~Zhang, V.~W. Zheng, and Z.~Yu, ``Modeling user activity preference by leveraging user spatial temporal characteristics in lbsns,'' \emph{IEEE Transactions on Systems, Man, and Cybernetics: Systems}, 2015.

\bibitem{friendship_2011}
E.~Cho, S.~A. Myers, and J.~Leskovec, ``Friendship and mobility: User movement in location-based social networks,'' in \emph{KDD '11}, 2011.

\bibitem{GMAN_2020}
C.~Zheng, X.~Fan, C.~Wang, and J.~Qi, ``Gman: A graph multi-attention network for traffic prediction,'' in \emph{AAAI}, 2020.

\bibitem{libcity_2021}
J.~Wang, J.~Jiang, W.~Jiang, C.~Li, and W.~X. Zhao, ``Libcity: An open library for traffic prediction,'' in \emph{Proceedings of the 29th International Conference on Advances in Geographic Information Systems}, 2021.

\bibitem{wang2023towards}
J.~Wang, J.~Jiang, W.~Jiang, C.~Han, and W.~X. Zhao, ``Towards efficient and comprehensive urban spatial-temporal prediction: A unified library and performance benchmark,'' \emph{arXiv preprint}, 2023.

\bibitem{shao2023exploring}
Z.~Shao, F.~Wang, Y.~Xu, W.~Wei, C.~Yu, Z.~Zhang, D.~Yao, G.~Jin, X.~Cao, G.~Cong \emph{et~al.}, ``Exploring progress in multivariate time series forecasting: Comprehensive benchmarking and heterogeneity analysis,'' \emph{arXiv preprint arXiv:2310.06119}, 2023.

\bibitem{wong2004modifiable}
D.~W. Wong, ``The modifiable areal unit problem (maup),'' in \emph{WorldMinds: geographical perspectives on 100 problems}.\hskip 1em plus 0.5em minus 0.4em\relax Springer, 2004, pp. 571--575.

\bibitem{openshaw1981modifiable}
S.~Openshaw, ``The modifiable areal unit problem,'' \emph{Quantitative geography: A British view}, pp. 60--69, 1981.

\bibitem{Zhang2016DNNbasedPM}
J.~Zhang, Y.~Zheng, D.~Qi, R.~Li, and X.~Yi, ``Dnn-based prediction model for spatio-temporal data,'' \emph{Proceedings of the 24th ACM SIGSPATIAL International Conference on Advances in Geographic Information Systems}, 2016.

\bibitem{li_traffic_2015}
Y.~Li, Y.~Zheng, H.~Zhang, and L.~Chen, ``Traffic prediction in a bike-sharing system,'' in \emph{Proceedings of the 23rd SIGSPATIAL International Conference on Advances in Geographic Information Systems}, 2015.

\bibitem{sthan_2022}
S.~Ling, Z.~Yu, S.~Cao, H.~Zhang, and S.~Hu, ``Sthan: Transportation demand forecasting with compound spatio-temporal relationships,'' \emph{ACM Trans. Knowl. Discov. Data}, 2022.

\bibitem{yuan2012segmentation}
N.~J. Yuan, Y.~Zheng, and X.~Xie, ``Segmentation of urban areas using road networks,'' Tech. Rep. MSR-TR-2012-65, July 2012.

\bibitem{pares2017fluid}
F.~Par{\'e}s, D.~G. Gasulla, A.~Vilalta, J.~Moreno, E.~Ayguad{\'e}, J.~Labarta, U.~Cort{\'e}s, and T.~Suzumura, ``Fluid communities: A competitive, scalable and diverse community detection algorithm,'' in \emph{International conference on complex networks and their applications}.\hskip 1em plus 0.5em minus 0.4em\relax Springer, 2017, pp. 229--240.

\bibitem{li2015traffic}
Y.~Li, Y.~Zheng, H.~Zhang, and L.~Chen, ``Traffic prediction in a bike-sharing system,'' in \emph{Proceedings of the 23rd SIGSPATIAL international conference on advances in geographic information systems}, 2015, pp. 1--10.

\bibitem{chen2016dynamic}
L.~Chen, D.~Zhang, L.~Wang, D.~Yang, X.~Ma, S.~Li, Z.~Wu, G.~Pan, T.-M.-T. Nguyen, and J.~Jakubowicz, ``Dynamic cluster-based over-demand prediction in bike sharing systems,'' in \emph{Proceedings of the 2016 ACM International Joint Conference on Pervasive and Ubiquitous Computing}, 2016, pp. 841--852.

\bibitem{1995Short}
M.~M. Hamed, H.~R. Al-Masaeid, and Z.~M.~B. Said, ``Short-term prediction of traffic volume in urban arterials,'' \emph{Journal of Transportation Engineering}, vol. 121, no.~3, pp. 249--254, 1995.

\bibitem{GeoMAN_2018}
Y.~Liang, S.~Ke, J.~Zhang, X.~Yi, and Y.~Zheng, ``Geoman: Multi-level attention networks for geo-sensory time series prediction,'' in \emph{IJCAI-18}, 2018.

\bibitem{geng2019spatiotemporal}
X.~Geng, Y.~Li, L.~Wang, L.~Zhang, Q.~Yang, J.~Ye, and Y.~Liu, ``Spatiotemporal multi-graph convolution network for ride-hailing demand forecasting,'' in \emph{Proceedings of the AAAI conference on artificial intelligence}, vol.~33, no.~01, 2019, pp. 3656--3663.

\bibitem{AGCRN_2020}
L.~Bai, L.~Yao, C.~Li, X.~Wang, and C.~Wang, ``Adaptive graph convolutional recurrent network for traffic forecasting,'' \emph{Advances in neural information processing systems}, vol.~33, pp. 17\,804--17\,815, 2020.

\bibitem{wu2020connecting}
Z.~Wu, S.~Pan, G.~Long, J.~Jiang, X.~Chang, and C.~Zhang, ``Connecting the dots: Multivariate time series forecasting with graph neural networks,'' in \emph{Proceedings of the 26th ACM SIGKDD international conference on knowledge discovery \& data mining}, 2020, pp. 753--763.

\bibitem{xgboost_2016}
T.~Chen and C.~Guestrin, ``Xgboost: A scalable tree boosting system,'' in \emph{KDD '16}, 2016.

\bibitem{chai_multi_graph_2018}
D.~Chai, L.~Wang, and Q.~Yang, ``Bike flow prediction with multi-graph convolutional networks,'' in \emph{Proceedings of the 26th ACM SIGSPATIAL International Conference on Advances in Geographic Information Systems}, 2018.

\bibitem{chebnet_2016}
M.~Defferrard, X.~Bresson, and P.~Vandergheynst, ``Convolutional neural networks on graphs with fast localized spectral filtering,'' in \emph{Proceedings of the 30th International Conference on Neural Information Processing Systems}, 2016.

\bibitem{velickovic2018graph}
P.~Veli{\v{c}}kovi{\'{c}}, G.~Cucurull, A.~Casanova, A.~Romero, P.~Li{\`{o}}, and Y.~Bengio, ``{Graph Attention Networks},'' \emph{ICLR}, 2018.

\end{thebibliography}

\end{document}